\pdfoutput=1

\documentclass[11pt]{article}
\usepackage[final]{acl}

\usepackage{makecell} 
\usepackage{times}
\usepackage{latexsym}
\usepackage{times}
\usepackage{latexsym}
\usepackage{amssymb}
\usepackage{graphicx}
\usepackage{amsmath}
\usepackage{multirow}
\usepackage{booktabs}
\usepackage{threeparttable}
\usepackage{longtable}  
\usepackage{makecell}
\usepackage{natbib}
\usepackage{colortbl}
\usepackage[most]{tcolorbox}
\usepackage{subfig}
\usepackage{tablefootnote}
\usepackage{CJKutf8}
\usepackage{url}
\usepackage{footmisc}
\usepackage{diagbox}
\usepackage{fontawesome5}

\usepackage[T1]{fontenc}

\usepackage[utf8]{inputenc}
\usepackage{microtype}

\usepackage{inconsolata}

\usepackage{graphicx}

\newcommand{\red}{\color{red}}

\newtcolorbox{myblock}[2][]{colback=blue!5!white, colframe=blue!75!black, fonttitle=\bfseries, title=#1, #2}

\title{T2R-bench: A Benchmark for Generating Article-Level Reports from \\ Real World Industrial Tables}

\author{
 \textbf{Jie Zhang\textsuperscript{1,*}},
 \textbf{Changzai Pan\textsuperscript{1,*}},
 \textbf{Kaiwen Wei\textsuperscript{2,*}},
 \textbf{Sishi Xiong\textsuperscript{1,*}},
\\
 \textbf{Yu Zhao\textsuperscript{1}},
 \textbf{Xiangyu Li\textsuperscript{1}},
 \textbf{Jiaxin Peng\textsuperscript{1}},
 \textbf{Xiaoyan Gu \textsuperscript{1}},
\\
 \textbf{Jian Yang\textsuperscript{3}},
 \textbf{Wenhan Chang\textsuperscript{1}},
 \textbf{Zhenhe Wu\textsuperscript{3}},
 \textbf{Jiang Zhong\textsuperscript{2}},
\\
 \textbf{Shuangyong Song\textsuperscript{1}},
 \textbf{Yongxiang Li\textsuperscript{1}},
 \textbf{Xuelong Li\textsuperscript{1,$\dagger$}} 
\\
 \textsuperscript{1} Institute of Artificial Intelligence (TeleAI), China Telecom,
 \\
 \textsuperscript{2} Chongqing University,
 \textsuperscript{3} Beihang University
}

\begin{document}
\maketitle

\renewcommand{\thefootnote}{\arabic{footnote}} 
\setcounter{footnote}{0}          
\let\thefootnote\relax\footnotetext{* These authors contributed equally to this work.}
\let\thefootnote\relax\footnotetext{$\dagger$ Corresponding author: xuelong\_li@ieee.org}

\begin{abstract}

Extensive research has been conducted to explore the capabilities of large language models (LLMs) in table reasoning. However, the essential task of transforming tables information into reports remains a significant challenge for industrial applications. This task is plagued by two critical issues: 1) the complexity and diversity of tables lead to suboptimal reasoning outcomes; and 2) existing table benchmarks lack the capacity to adequately assess the practical application of this task. To fill this gap, we propose the \textbf{table-to-report} task and construct a bilingual benchmark named \textbf{T2R-bench}, where the key information flow from the tables to the reports for this task. The benchmark comprises 457 industrial tables, all derived from real-world scenarios and encompassing 19 industry domains as well as 4 types of industrial tables. Furthermore, we propose an evaluation criteria to fairly measure the quality of report generation. 
The experiments on 25 widely-used LLMs reveal that even state-of-the-art models like Deepseek-R1 only achieves  performance with \text{62.71\%} overall score, indicating that LLMs still have room for improvement on T2R-bench. Data can be found

\end{abstract}

\noindent 
\textbf{Data:} \\
{\small
\href{https://huggingface.co/datasets/Tele-AI/TeleTableBench}{https://huggingface.co/datasets/Tele-AI/TeleTableBench}
\href{https://github.com/Tele-AI/TeleTableBench}{https://github.com/Tele-AI/TeleTableBench} \\
}

\section{Introduction}
\begin{figure}[h]
    \centering 
    \includegraphics[width=0.95\linewidth]{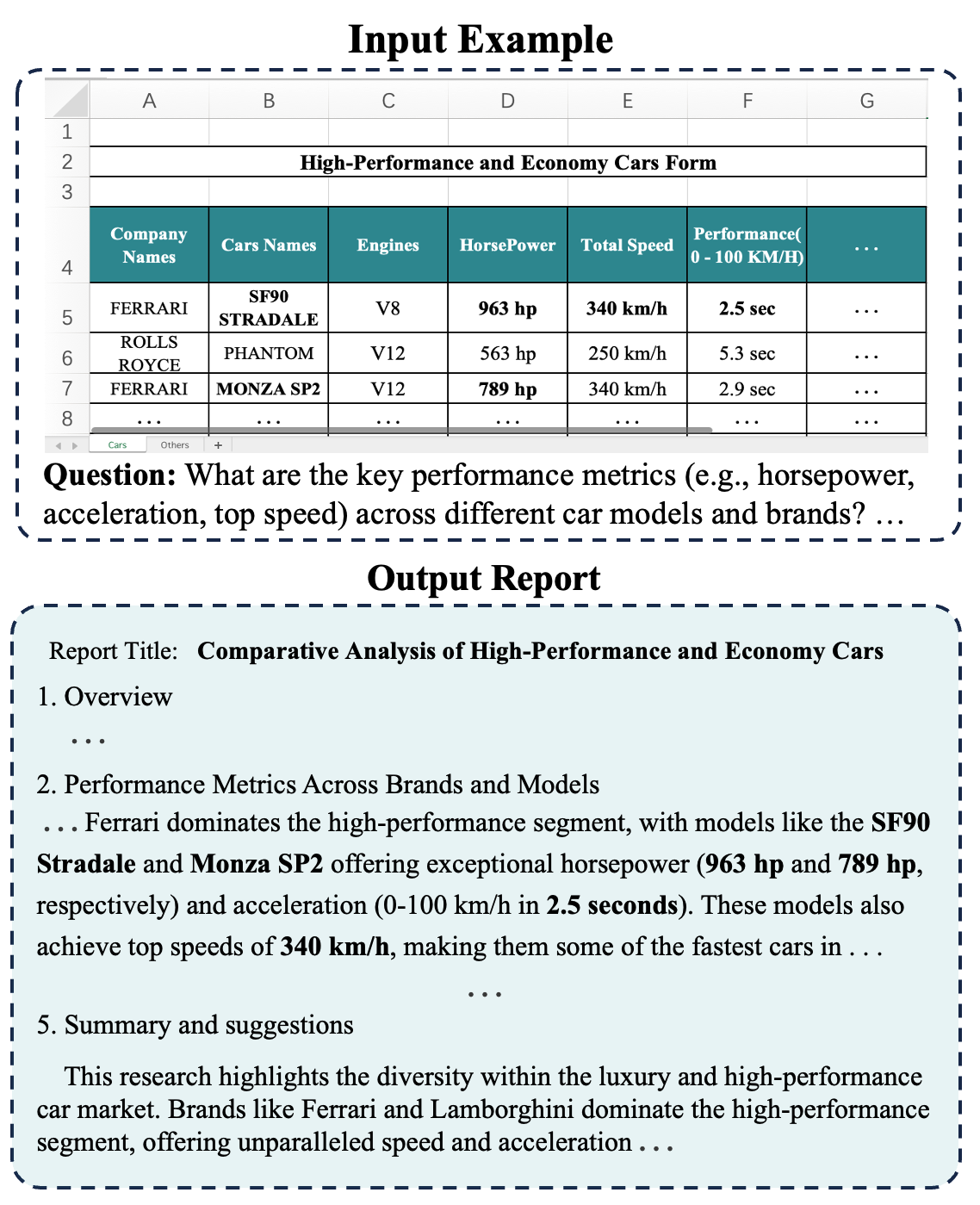} 
    \caption{The illustration of table-to-report task. The goal of this task is to analyze numerical data from table to generate comprehensive, coherent and accurate report, including descriptions, analysis and conclusions.}
    \label{Fig:intro}
\end{figure}

The rapid development of large language models (LLMs) has significantly advanced research progress in table reasoning \citep{Survey}. Traditional research has primarily focused on tasks such as table-to-text generation \citep{ToTTo}, table question answering \citep{WikiTableQuestion, InfiAgent-DABench, xiong2025teleai, xiong2025tablereasoner}, fact verification \citep{chen2019tabfact}, text2sql \citep{li2024scalable, wu2025uniting,MR-SQL} and table data analysis \citep{chen2023largelanguagemodelsfew1shot}. 

However, the automation of report generation from tables is far more widely adopted in industrial applications\footnote{Typical industrial table applications include Microsoft Power BI, SAP BusinessObjects, IBM Cognos, MicroStrategy, Smartbi, etc.}, such as industrial data analysis systems \cite{ma-etal-2023-insightpilot}, business intelligence (BI), table analysis tools and enterprise reporting tools. Notably, systematic research in this field remains largely unexplored and urgently requires further in-depth investigation. In addition, industrial tables commonly exhibit high complexity and diversity, creating a significant gap between existing academic benchmarks and industrial demands, which poses new challenges for related research.

In light of the above considerations, we propose \textbf{table-to-report}, a novel task designed to convert structured tabular data into natural language reports, aiming to present data, trends, and insights to enhance table information flow efficiency \cite{aiflow-2024} as illustrated in Figure~\ref{Fig:intro}.

As an emerging task, it still faces several key challenges: (1) \textbf{Lack of high-quality benchmark}: current table benchmarks, such as Text2Analysis \citep{Text2Analysis}, TableBench \citep{wu2024tablebench} and MiMoTable \citep{li2024mimotable} primarily evaluate LLMs on table question answering tasks, each with a distinct focus. However, these benchmarks are not designed to assess table-to-report task. Besides, the table datasets used in most benchmarks predominantly consist of open-source academic data, failing to fully capture the main features and types of industrial tabular data, such as multiple tables, complex structure, and extremely large-size tables. The data volume and scale remain significantly constrained for extremely large-size tables \citep{SUC}. (2) \textbf{Lack of targeted evaluation criteria}: existing criteria like BLEU \citep{Papineni2002BleuAM} and ROUGE \citep{lin2004rouge} designed for summarization tasks, are unsuitable for table-to-report task due to non-unique gold standards. While general LLM-as-a-judge \citep{li2024llmasajudge} method performs excel in text quality assessment,it neglects to evaluate numerical accuracy and table topic coverage, therefore limiting its applicability. 

To address the aforementioned issues, we introduce \textbf{T2R-bench}, a high-quality benchmark designed to evaluate the reasoning capabilities of LLMs in the table-to-report task. T2R-bench encompasses Chinese and English tables from real-world industrial scenarios, covering 6 domains and 19 secondary industry categories. Compared to existing table benchmarks, as shown in Table~\ref{Tab:table_type}, our benchmark features a comprehensive and diverse collection of single tables, multiple tables, complex structured tables, and extremely large-size tables, enhancing the benchmark's practicality and challenge. We also craft the designed approaches for table question annotation and report reference annotation. Furthermore, we develop an evaluation system that incorporates three criteria of numerical accuracy, information coverage, and general quality to comprehensively assess report quality. In the experiment, we select 25 widely-used methods for evaluation, the results demonstrate that strongest models struggle to achieve satisfactory performance on the table-to-report task. 

Our contributions are summarized as follows:

\begin{table}[!ht]
\centering
\resizebox{0.9\linewidth}{!}{
\begin{tabular}{c c c c c c}
    \toprule 
    Task and Benchmark & \makecell[c]{Multiple\\Table} & \makecell[c]{Complex\\Structure\\Table} & \makecell[c]{Extremely\\Large-Size\\Table} & \makecell[c]{Answer\\Lengths} \\
    \midrule 
    \multicolumn{3}{l}{\textbf{TableQA}}\\
    WikiSQL \cite{zhong2017seq2sqlgeneratingstructuredqueries} & $\times$ & $\times$ & $\times$ & 1.9\\
    WTQ \cite{pasupat-liang-2015-compositional} & $\times$ & $\times$ & $\times$ & 10.39\\
    TAT-QA \cite{zhu-etal-2021-tat} & $\times$ & $\times$ & $\times$ & 20.3\\
    FeTaQA \cite{nan2021fetaqafreeformtablequestion} & $\times$ & $\times$ & $\times$ & 18.9\\
    AIT \cite{katsis2021aitqaquestionansweringdataset} & $\times$ & $\times$ & $\times$ & 1.1\\
    TabFact \cite{chen2020tabfactlargescaledatasettablebased} & $\times$ & $\times$ & $\times$ & 18.3\\
    TableBench \cite{wu2024tablebench} & $\times$ & $\times$ & $\times$ & 8.5\\
    HiTab \cite{cheng2022hitabhierarchicaltabledataset} & $\times$ & \checkmark & $\times$ & 12.9\\
    DataBench \cite{grijalba2024question} & $\times$ & $\times$ & \checkmark & 3.2\\
    Mimo \cite{li2024mimotable} & \checkmark & \checkmark & $\times$ & 44.2\\
    Spider \cite{yu-etal-2018-spider} & \checkmark & $\times$ & \checkmark & 35.5\\
    \midrule 
    \multicolumn{3}{l}{\textbf{Table2Text}}\\
    ToTTo \cite{ToTTo} & $\times$ & $\times$ & \checkmark & 17.4\\
    DAE-val \cite{hu2024infiagentdabenchevaluatingagentsdata} & $\times$ & $\times$ & \checkmark & 3.6\\
    DataTales \cite{yang-etal-2024-datatales} & $\times$ & $\times$ & \checkmark & 108.0\\
    Text2Analysis \cite{Text2Analysis} & $\times$ & $\times$ & $\times$ & /\\
    \midrule 
    \multicolumn{3}{l}{\textbf{Table2Report}}\\
    T2R-Bench (ours) & \checkmark & \checkmark & \checkmark & 950.2\\
    \bottomrule 
\end{tabular}}
\caption{Comparison with existing datasets on table types and answer lengths. Since Text2Analysis benchmark dose not provide the publicly accessible download links, the average length could not be calculated.}
\label{Tab:table_type}
\end{table}

{
\begin{itemize}
    \item We introduce \textbf{T2R-bench}, the first real world industrial benchmark for the table-to-report task. It encompasses 457 real-world tables across 19 domains, covering 4 industrially relevant types, including single tables, multiple tables, complex structured tables, and extremely large-size tables.

    \item We propose an evaluation system for table-to-report generation, incorporating 3 carefully designed criteria to assess report accuracy and reliability. Extensive validation demonstrates that the evaluation system achieves strong alignment with human judgment.


    \item We evaluate the ability of 25 strong methods on T2R-Bench. The experiments show that the best performed model Deepseek-R1 achieves only 62.71\% overall score, which suggests great challenges in satisfying real-world table-based report generation needs.

\end{itemize}
}

%

\section{Related Work}
\textbf{Tabular Benchmarks.} With the development of deep learning~\cite{DBLP:conf/acl/WeiSZZGJ20, DBLP:conf/acl/WeiYJSZZLZLZ23, DBLP:journals/tkde/WeiSZJZLG23,xing2025llmsr, wang2025when,zhao2025enhancing,wu2025table,wang-etal-2024-boosting-llm, wang2025beyond, dai2025secure, dai2025captions}, recent advances in table reasoning research have driven the development of diverse benchmarks covering TableQA, Table2Text, and advanced data analysis tasks, incorporating various table types including large-size tables, multiple tables, and complex structures. TableQA benchmarks  \cite{zhong2017seq2sqlgeneratingstructuredqueries,chen2020tabfactlargescaledatasettablebased,nan2021fetaqafreeformtablequestion,oses-grijalba-etal-2024-question} dominate the landscape, with TableBench \cite{wu2024tablebench} emerging as a representative benchmark that captures real-world tabular reasoning challenges. For Table2Text tasks~\cite{lebret2016neuraltextgenerationstructured}, ToTTo~\cite{ToTTo} constructs table-description pairs from Wikipedia snippets, while DATATALES~\cite{yang-etal-2024-datatales} generates financial narratives from tabular data. Advanced analysis benchmarks like DAEval \cite{hu2024infiagentdabenchevaluatingagentsdata} and Text2Analysis \cite{Text2Analysis} focus on programmatic table manipulation. However, as evidenced in Table~\ref{Tab:table_type}, current solutions remain limited in their coverage of diverse table types (including large-scale, multi-table, and complex layouts) and are constrained to sentence-level outputs that fail to meet industrial requirements for comprehensive report generation.

Recent research has placed growing emphasis on complex table structure understanding \cite{cheng2022hitabhierarchicaltabledataset,katsis2021aitqaquestionansweringdataset,tang-etal-2024-struc, mathur-etal-2024-matsa}, yielding specialized benchmarks like MiMoTable \cite{li2024mimotable} for multidimensional spreadsheets, DataBench \cite{grijalba2024question} for containing a limited number of large-size tables, and SPREADSHEETBENCH \cite{ma2024spreadsheetbenchchallengingrealworld} for multiple tables manipulation. However, these works focused on TableQA and manipulation tasks, overlooking comprehensive report generation needs. \\
\textbf{Text quality Evaluation}. Established metrics like ROUGE \cite{lin2004rouge}, BLEU \cite{papineni2002bleu}, and BERTScore \cite{zhang2020bertscoreevaluatingtextgeneration} have been widely adopted, complemented by emerging LLM-as-judge approaches \cite{li2024llmasajudge}. For Table2Text tasks, Text2Analysis employs code generation metrics, while \cite{wiseman2017challengesdatatodocumentgeneration} designs three new dataset-adapted evaluation metrics for text generation. ToTTo~\cite{li2024mimotable} adapts PARENT \cite{dhingra2019handlingdivergentreferencetexts} alongside BLEU. DATATALES introduces domain-specific criteria including factual accuracy, insightfulness, and stylistic quality, demonstrating the necessity for task-aligned evaluation frameworks. However, those methods typically neglect to evaluate numerical accuracy and table topic coverage~\cite{DBLP:conf/iui/SzymanskiZEL0M25}, hindering the evaluation applicability.

\section{Construction of T2R-bench}
\label{Sec:bench construction}

\begin{figure*}[htbp]
    \centering 
    \includegraphics[width=0.99\textwidth]{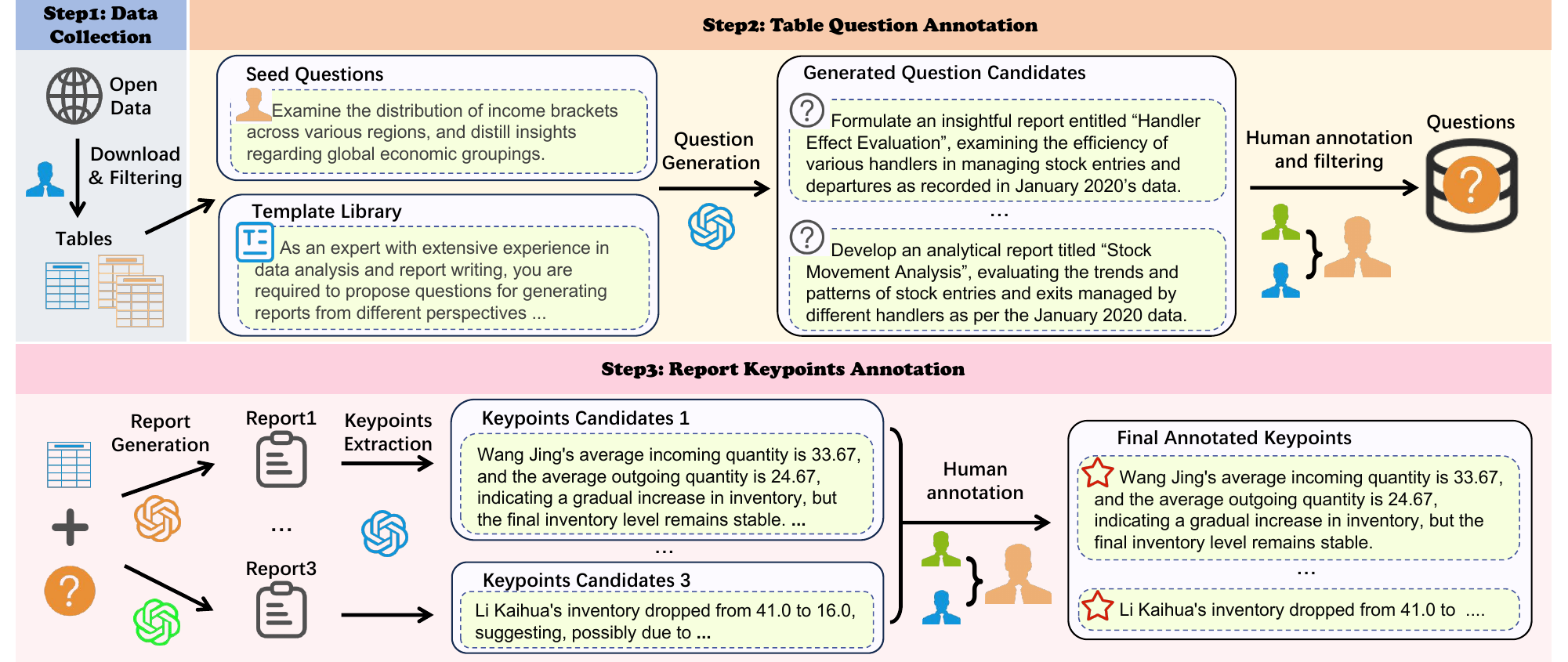} 
    \caption{An overview of the construction pipeline for T2R-bench.}
    \label{Fig:bench_pipeline}
\end{figure*}

Table-to-report is the task of automatically converting a structured table $T$ into a fluent article-level report $R$.
To evaluate existing approaches, we introduce T2R-bench, whose construction pipeline consists of three key components: table data collection, table question annotation, and report reference annotation, as detailed in Figure~\ref{Fig:bench_pipeline}.

\subsection{Table Data Collection}

The tables of T2R-bench are collected from publicly available internet resources. The primary sources encompass municipal open data platforms, the National Bureau of Statistics, industrial association official websites and open-source tabular datasets (refer to Appendix~\ref{Appd:source} for details). We collect tables to cover as many real-world scenarios as possible, including single table with individual and multiple sheets, multiple tables, extremely large-size tables, tables with simple and complex header structures. 

Specifically, we leverage a two-stage selecting method. Firstly, tables are pre-screened based on industry-specific topics to ensure domain relevance. Subsequently, to ensure each table has sufficient information density for statistical analysis,
we remove files with obvious garbled text or cell blank values exceeding 60\%. To ensure the quality and legality of the collected tables, we manually review each table and anonymize any potential private and sensitive information. 
Ultimately, we collect 252 Chinese tables and 205 English tables across 6 distinct domains and 19 secondary classes based on topics to fit diverse industrial fields. 


\subsection{Table Question Annotation}



We adopt a semi-automatic heuristic method to efficiently generate diverse and high-quality questions. The specific steps are shown as follows:
\\
\textbf{Seed Question and Prompt Preparing}. To improve the precision and relevance of the generated questions, we employ 24 annotators with expertise in data analysis and report writing in diverse domains (see Appendix~\ref{Appd:annotation} for annotator qualifications). They carefully curate 10 seed questions, and meticulously design the prompt template library with 5 diverse prompt templates (prompt templates and seed questions are provided in Appendix~\ref{Appd:question_gernation_prompt}).
\\
\textbf{Self-Instruct to Generate Questions}. We employ self-instruct\citep{wang-etal-2023-self-instruct} by using GPT-4o to efficiently generate a pool of questions. Two prompt templates are randomly selected from the prompt template library for each table. Each template incorporates 2-5 seed questions as in-context demonstrations, with instructions to generate 3 relevant questions.
\\
\textbf{Human Annotation and Filtering}. 
We randomly assign each question to two annotators, whose selection criteria and qualifications are detailed in Appendix~\ref{Appd:annotation}. 
Annotators evaluate question candidates based on three criteria: 1) tabular answer ability, where questions must be answerable solely using table data without external knowledge; 2) focused conclusions, where questions should target single analytical dimensions for definitive conclusions; and 3) complementary uniqueness, where questions from the same table must address distinct aspects.
In cases where the evaluation results of the two annotators are inconsistent, the results will be handed over to a third senior annotator with extensive domain expertise and experience for the final judgment (For detailed annotation procedure, please refer to Appendix~\ref{Appd:annotation_procedure}). Through this rigorous quality assurance procedure, we obtain 910 high-quality, comprehensive questions.

\subsection{Report Reference Annotation}
\label{Sec:report_reference}

Unlike summarization tasks, which often yield a single optimal summary, table-to-report tasks exhibit significant variability due to differences in expression, stylistic preferences, structural choices among annotators, and the inherent complexity of tabular data. Consequently, using entire reports as reference standards proves impractical.

To this end, we observe that professionally authored reports on the same tabular content and report topic consistently share core elements, including central viewpoints, analytical conclusions, recommendations, critical supporting data, despite variations in phrasing or presentation.
This consistency motivates our introduction of report keypoints: distilled representations of a report’s essential content, encompassing its analytical backbone and evidentiary support (See Figure~\ref{Fig:bench_pipeline} for keypoint examples). These invariant keypoints provide a robust basis for evaluating generated reports.


Based on this finding, we design the report reference annotation process, which consists: 1) \textbf{Report Generation}. We leverage three distinct LLMs to generate different reports for each <table, report question> pair, resulting in three distinct reports (see Appendix~\ref{Appd:prompt_report} for the prompt template).
2) \textbf{Keypoints Extraction}. Then, we prompt GPT-4o to distill the most crucial information from each report, extracting 5-10 keypoints, resulting three groups of keypoints for each <table, question> pair (see Appendix~\ref{Appd:report_core_points_extraction} for the prompt template). 3) \textbf{Human Annotation}.  Mirroring the question annotation procedure, we implement a rigorous dual-annotator verification protocol for key point refinement, with discrepancies resolved by senior annotators with data analysis and domain-specific report writing experience. Please see Appendix~\ref{Appd:annotation} and \ref{Appd:key_points_annotation} for full qualifications and annotation procedure details.



\subsection{Dataset Statistics}
\label{Sec:dataset_statistics}
Through the construction process, T2R-bench comprises 910 high-quality questions originating from 457 unique tables, along with 4,320 annotated report keypoints. These meticulously annotated keypoints of the report will serve as the gold reference to evaluate report in Section~\ref{Sec:infomation_coverage}.
\captionsetup[subfloat]{font=scriptsize}
\begin{figure}[t!]
    \captionsetup[subfigure]{captionskip=-10pt, nearskip=-5pt}
    \centering
	\begin{minipage}{0.65\linewidth}
        \centering 
        \subfloat[]{
        \includegraphics[width=\textwidth]{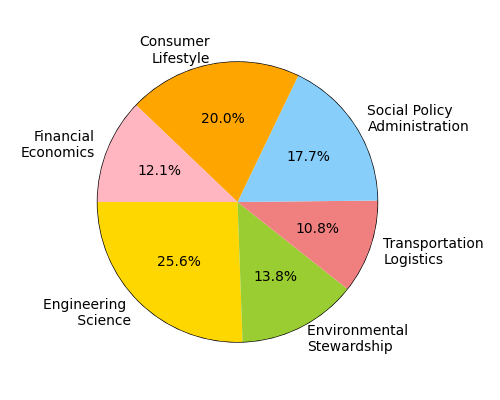}
        \label{Fig:domain_distribution}
        }
        \vspace{5pt} 
        \end{minipage}
    \captionsetup[subfigure]{captionskip=0pt, nearskip=0.0pt}
    \begin{minipage}{0.30\linewidth}
        \centering
	\begin{minipage}{\linewidth}
        \centering 
        \subfloat[]{
        \includegraphics[width=\textwidth]{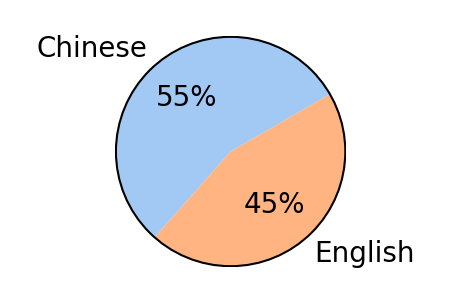} 
        }
        \end{minipage}
        \begin{minipage}{\linewidth}
        \centering 
        \subfloat[]{
        \includegraphics[width=\textwidth]{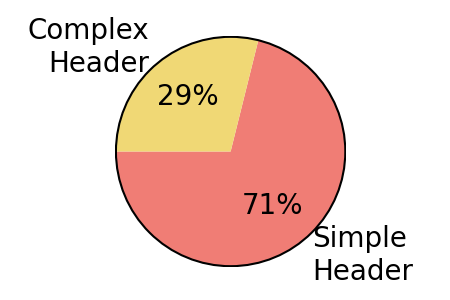} 
        }
        \end{minipage}
    \end{minipage}

        \begin{minipage}{0.48\linewidth}
        \centering 
        \subfloat[]{
        \includegraphics[width=\textwidth]{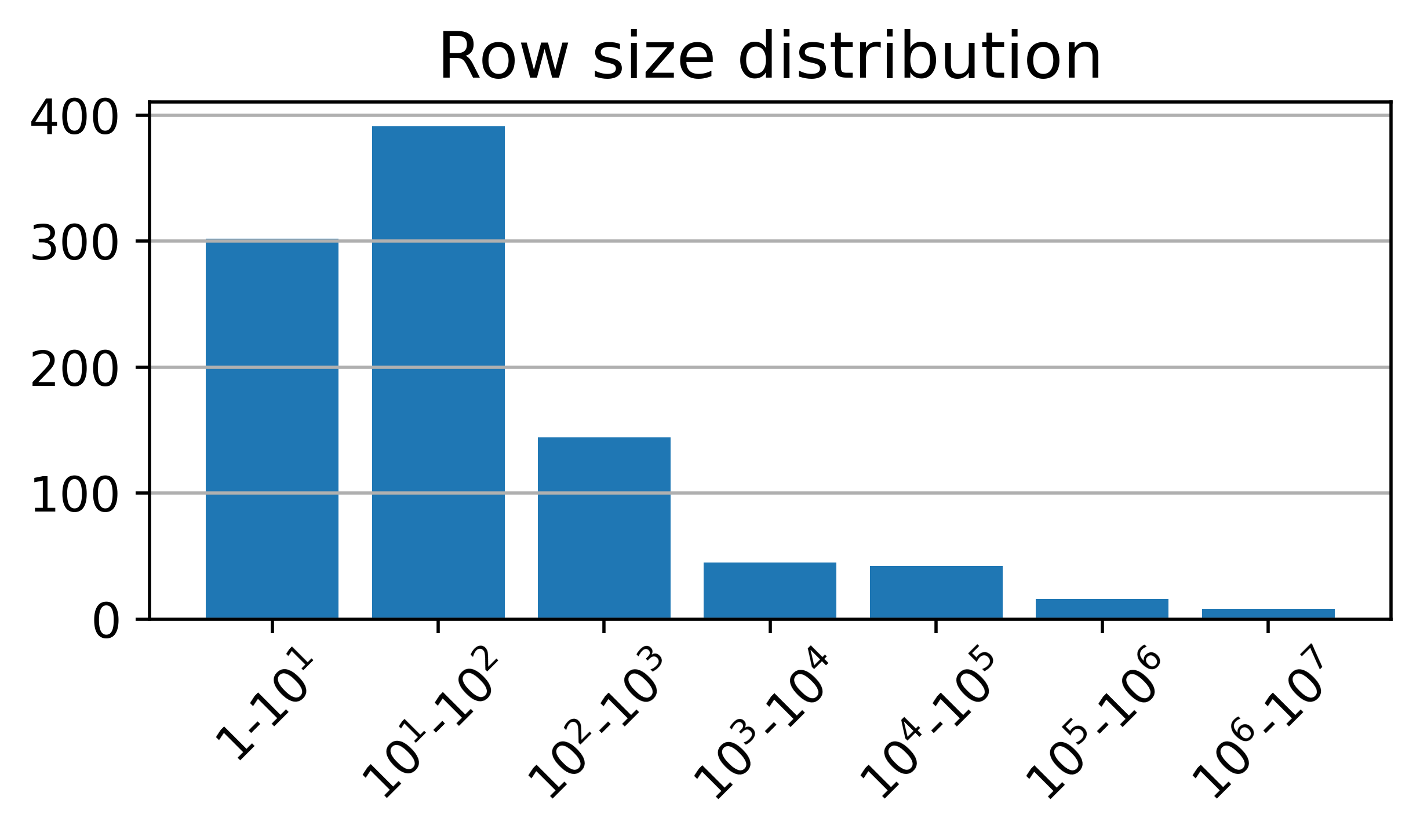} 
        }
        \end{minipage}
        \begin{minipage}{0.48\linewidth}
            \centering 
            \subfloat[]{
            \includegraphics[width=\textwidth]{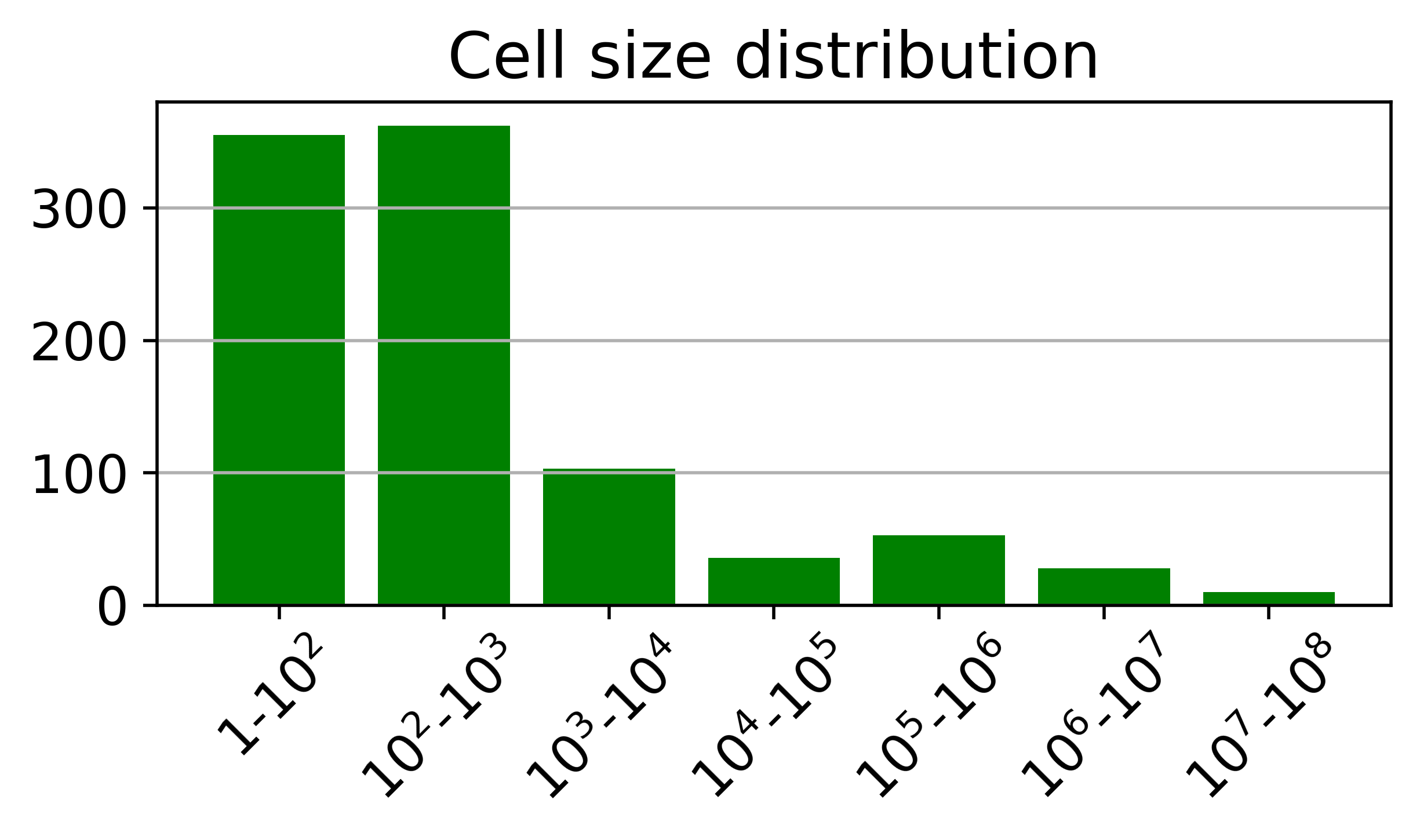} 
            }
            \end{minipage}
        
        \begin{minipage}{0.48\linewidth}
        \centering 
        \subfloat[]{
        \includegraphics[width=\textwidth]{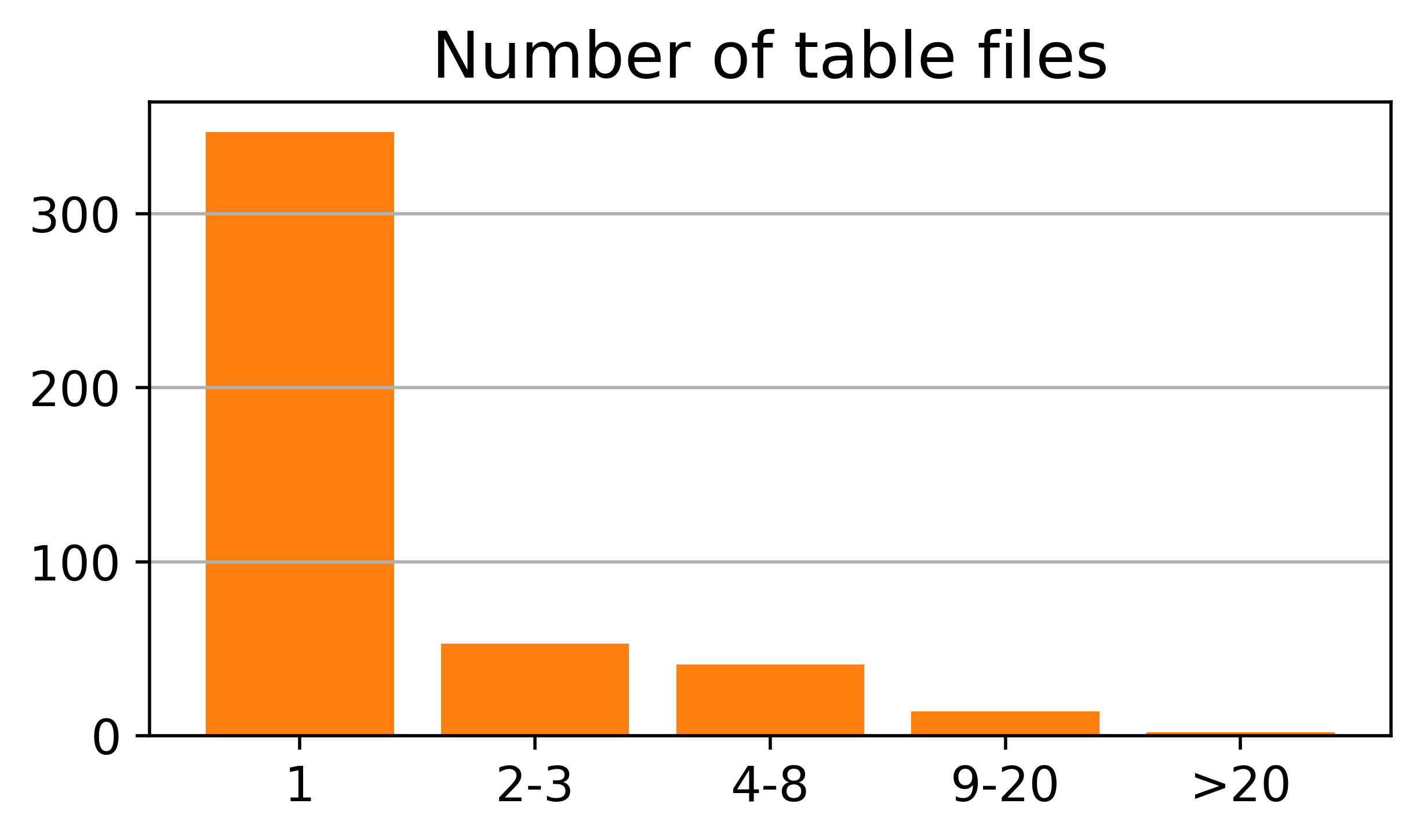} 
        }
        \end{minipage}
        \begin{minipage}{0.48\linewidth}
        \centering 
        \subfloat[][]{
        \includegraphics[width=\textwidth]{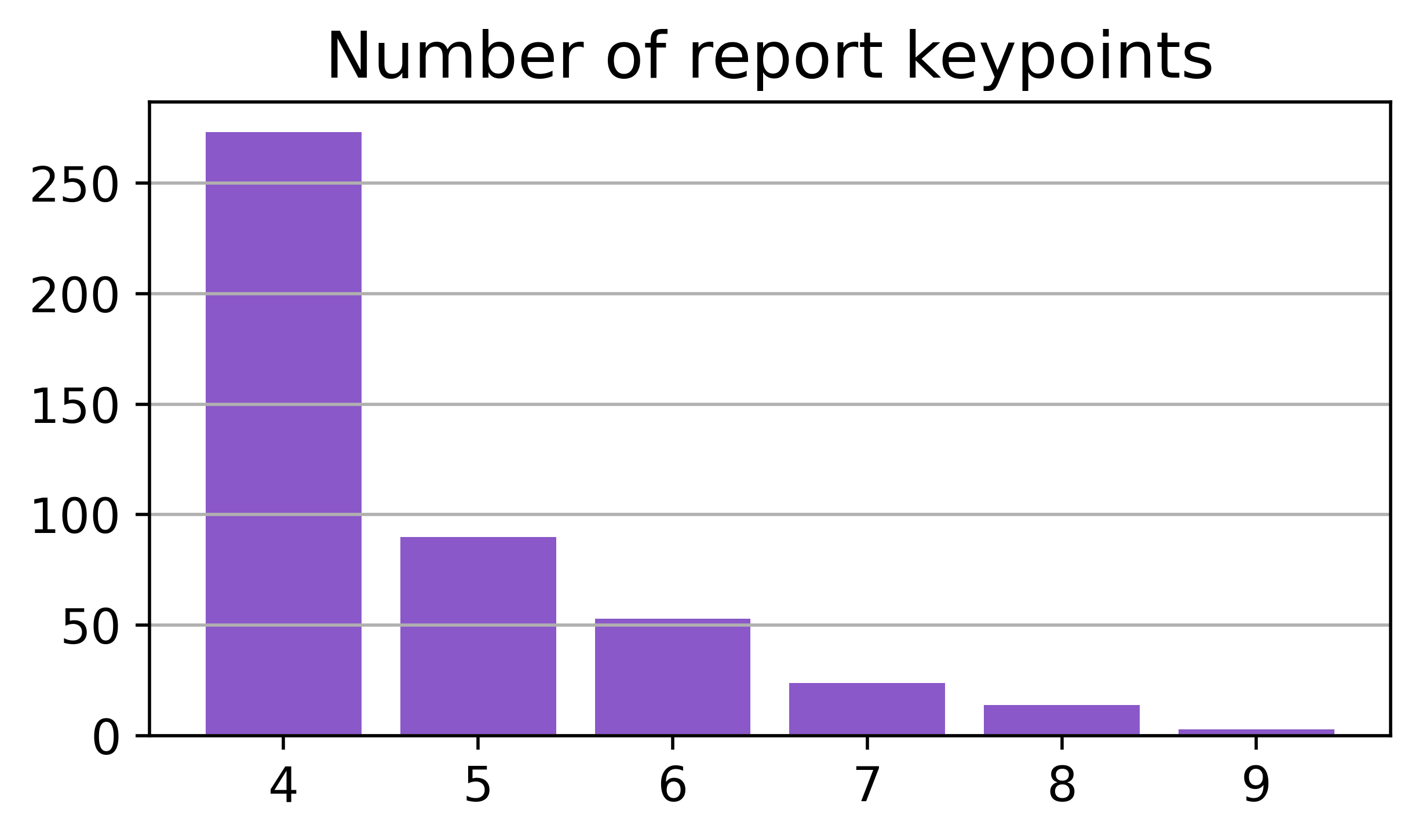} 
        }
        \end{minipage}
\caption{Distribution of different types of tables in T2R-bench. (a) Domain distribution. (b) Proportion of Chinese and English tables. (c) Proportion of complex header tables. (d-e) The row and cell size distribution for all tables in T2R-bench. (f) Number of table files for single tables and multiple tables. (g) Distribution of report reference keypoints for each report question.}
\label{Fig:data_distribution}
\end{figure}
\\
\textbf{Table Statistics.} Table~\ref{Tab:key_stat} and Figure~\ref{Fig:data_distribution} show the key statistics and distribution of tables in T2R-bench. Specifically, the global statistics reveal that T2R-bench contains over 8.3\% extremely large-size tables containing more than 50K cells; 28.9\% complex structured tables with hierarchical indexing, merged cells, and non-uniform cell structures; and 23.6\% multiple tables comprising interdependent files or sheets. A key feature distinguishing our benchmark is its substantial number of extremely large-size tables.
\\
\textbf{Domain Distribution.} As shown in Figure~\ref{Fig:domain_distribution}, T2R-bench covers six main industry domains, which can be further divided into 19 more specific sub-domains, including engineering, manufacturing, finance, education, healthcare, telecommunications, transportation (detailed sub-categories refers to Table~\ref{Tab:domains} of Appendix~\ref{Appd:domain_distribution}), ensuring that abundant types of tables in the dataset encompass as many real-world scenarios as possible.
\\
\textbf{Questions and Report Keypoints.}  Following the human annotation process, T2R-bench comprises a total of 910 questions. Notably, the number of questions is pruned from an initial range of 3.00 to 1.99 per table during the expert annotation phase. 
As illustrated in Figure~\ref{Tab:key_stat}, the number of report keypoints per question is reduced to an average of 4.75 after expert verification and filtering. These rigorous annotation and verification processes enhance the quality of the benchmark.

\section{Evaluation Criteria}
\label{Sec:criteria}
To address the challenges encountered in automated evaluation for the table-to-report task, we propose a comprehensive evaluation system from three aspects: numerical accuracy, information coverage, and general quality.


\begin{table}[t]
\centering
   \resizebox{0.8\linewidth}{!}{
    \begin{tabular}{ll}
    \toprule
    \textbf{Property} & \textbf{Value} \\ 
    \midrule
    Number of Tables & 457 \\ 
    Avg Table Files or Sheets for Multi-Tables & 5.04 \\
    Avg Rows per Table & 30,183 \\
    Avg Cells per Table & 490,308 \\
    Number of Extremely Large-size Tables & 38 \\
    Avg Rows for Extremely Large-size Tables & 721,882 \\
    Avg Cells for Extremely Large-size Tables & 11,895,814 \\
    Number of Questions & 910 \\ 
    Avg Questions per Table & 1.99\\
    Avg Report Reference Keypoints per Question & 4.75\\
    \bottomrule
    \end{tabular}}
    \caption{Key Statistics of T2R-bench.}
    \label{Tab:key_stat}
\end{table}

\subsection{Numerical Accuracy Criterion}

Generated reports frequently incorporate numerical values, some directly extracted from source tables and others derived through data synthesis (e.g., aggregations like column averages). To ensure the fidelity of such numerical claims, we propose the Numerical Accuracy Criterion (NAC), a self-consistency mechanism for validating numerical facts against their tabular sources.

Specifically, we first segment sentences in the target report using NLTK\footnote{https://www.nltk.org/} (for English) and Jieba\footnote{https://github.com/fxsjy/jieba} (for Chinese). We then apply regular expressions to extract clusters of sentences containing numerical statements (integers or floating-point numbers). For each cluster, we generate corresponding verification questions, treating the extracted numerical values as ground-truth answers (see Appendix~\ref{Appd:NAC} for prompt).

To resolve these questions robustly, we employ three specialized code-generation LLMs (i.e., Qwen2.5-32B-Coder-Instruct, Deepseek-Coder, and CodeLlama-70B-Instruct), capable of interpreting and executing numerical operations (see Appendix~\ref{Appd:NAC} for details). NAC enforces consensus by requiring agreement from at least two models; discordant results (including execution failures) are discarded to minimize noise.
The final NAC score is computed by systematically comparing the validated solutions against the original numerical assertions in each sentence cluster.

\subsection{Information Coverage Criterion}
\label{Sec:infomation_coverage}

To address the challenges of incomplete coverage and irrelevant content in LLM-generated reports, we propose the Information Coverage Criterion (ICC), a quantitative measure of semantic alignment between generated reports and reference keypoints. Inspired by the successful application of mutual information (MI) in machine translation for evaluating alignment quality, ICC assesses how effectively a report preserves essential information from the source table.

Specifically, for each generated report, we define $K = \{k_1, k_2, \dots, k_M\}$ as the set of annotated keypoints, where $M$ represents the total keypoint number. 
Then, the generated report is segmented into multiple sentence clusters $S=\{s_1, s_2, \dots, s_N\}$ by NLTK toolkit (English reports) and Jieba toolkit (Chinese reports).
After that, we construct a semantic similarity matrix $S$, where each element $S_{ij}$ represents the semantic similarity of keypoints-sentence pair($k_i$, $s_j$) calculated by BERTScore\cite{zhang2020bertscoreevaluatingtextgeneration}:
\begin{equation} 
    S_{ij} = BERTScore(k_i, s_j) \nonumber
\end{equation}

Given the similarity matrix $S$, the ICC is defined as normalized MI:
\begin{equation}
    ICC = \frac{\sum_{i=1}^M \sum_{j=1}^N P(k_i, s_j) \log\frac{P(k_i, s_j)}{P(k_i)P(s_j)}}{-\sum_{i=1}^M P(k_i)\log P(k_i)}
    \label{eq:icc}
\end{equation}
where the joint and marginal probabilities are derived from similarity matrix S as follows:
\begin{eqnarray}
    P(k_i, s_j) &=& \frac{S(k_i, s_j)}{\sum_{i=1}^M\sum_{j=1}^N S(k_i, s_j)} \nonumber\\
    P(k_i) &=& \frac{\sum_{j=1}^N S(k_i, s_j)}{\sum_{i=1}^M\sum_{j=1}^N S(k_i, s_j)} \nonumber\\
    P(s_j) &=& \frac{\sum_{i=1}^M S(k_i, s_j)}{\sum_{i=1}^M\sum_{j=1}^N S(k_i, s_j)} \nonumber
\end{eqnarray}

Eq.~\eqref{eq:icc} provides an information-theoretic measure scaled to [0,1] by dividing the keypoint entropy $H(K)$, enabling consistent comparison across reports with varying numbers of keypoints. The final evaluation aggregates ICC scores across all reports, with higher values indicating better preservation of critical information in the generated outputs.

\subsection{General Evaluation Criterion}

Inspired by evaluation methodologies for long-context generation~\cite{DBLP:journals/corr/abs-2410-16848}, we propose the General Evaluation Criterion (GEC) to holistically assess report quality using LLMs as judges. GEC focuses on five key dimensions that most effectively discriminate report quality: reasoning depth, human-like style, practicality, content completeness and logical coherence. The final GEC score is computed as the average across these dimensions. Detailed evaluation criteria and prompts are provided in Appendix~\ref{Appd:GEC_prompt}.



\begin{table*}[ht!]
    \centering
    \resizebox{\textwidth}{!}{
    \begin{tabular}{l| cccc c ccc c ccc c ccc c ccc}
        \toprule
        \multirow{2}{*}{\textbf{Model}}  & \multicolumn{4}{c}{\textbf{Overall}} & & \multicolumn{3}{c}{\textbf{Single}} & & \multicolumn{3}{c}{\textbf{Multiple}} & & \multicolumn{3}{c}{\textbf{Complex Structure}} & & \multicolumn{3}{c}{\textbf{Extremely Large-Size}}\\
        & \textbf{NAC} & \textbf{ICC} & \textbf{GEC} & \textbf{AVG} & & 
        \textbf{NAC} & \textbf{ICC} & \textbf{GEC} & &
        \textbf{NAC} & \textbf{ICC} & \textbf{GEC} & &
        \textbf{NAC} & \textbf{ICC} & \textbf{GEC} & &
        \textbf{NAC} & \textbf{ICC} & \textbf{GEC}\\
        \midrule
        \multicolumn{3}{l}{\textbf{Open-Source Models}}\\
        TableGPT2-7B \citep{su2024tablegpt2largemultimodalmodel} & 34.24 & 25.70 & 81.28 & 47.07 && 49.54 & 35.91 & 83.09 && 40.99 & 31.29 & 81.76 && 30.12 & 27.40 & 79.87 && 16.33 & 8.20 & 80.42\\
        Qwen1.5-14B-Chat \citep{qwen} & 36.03 & 26.29 & 83.83 & 48.72 && 50.67 & 46.62 & 82.61 && 38.31 & 27.06 & 82.72 && 40.02 & 25.67 & 85.17 && 15.12 & 5.81 & 84.82\\ 
        Qwen2.5-72B-Instruct \citep{qwen2025qwen25technicalreport} & 47.82 & 42.28 & 88.68 & 59.59 && 67.29 & 58.23 & 87.82 && 54.15 & 46.23 & \underline{89.65} && 43.58 & 47.40 & \textbf{90.42} && 26.18 & 17.24 & 86.84\\ 
        Qwen2-72B \citep{qwen2} & 44.64 & 33.76 & 87.76 & 55.39 && 66.23 & 50.36 & 88.55 && 46.96 & 39.35 & 88.71 && 39.92 & 29.93 & 88.53 && 25.47 & 15.40 & 85.25\\ 
        Qwen2.5-32B \citep{qwen2025qwen25technicalreport} & 42.91 & 35.85 & 84.54 & 54.43 && 54.36 & 44.45 & 79.45 && 50.03 & 46.53 & 86.82 && 45.84 & 40.21 & 88.64 && 21.41 & 12.21 & 83.26\\ 
        Qwen2.5-Coder-32B-Instruct \citep{hui2024qwen25codertechnicalreport} & 43.82 & 32.33 & 86.25 & 54.13 && 61.97 & 46.36 & 86.18 && 44.57 & 40.82 & 86.24 && 46.23 & 28.53 & 86.13 && 22.52 & 13.62 & 86.43\\ 
        Qwen3-30B-A3B \citep{qwen2025qwen25technicalreport} & 49.46 & 42.27 & 88.02 & 59.90 && 70.32 & 56.46 & \underline{90.35} && 54.35 & 47.35 & 88.36 && 47.82 & 45.65 & 87.02 && 25.36 & 19.63 & 86.24\\ 
        Qwen3-32B \citep{qwen2025qwen25technicalreport} & \underline{53.01} & \underline{45.01} & 89.12 & \underline{61.37} && 73.21 & 59.34 & \textbf{91.21} && \underline{58.24} & 50.53 & \textbf{90.23} && 51.24 & 48.82 & 88.53 && \underline{29.35} & \textbf{21.25} & 88.82\\ 
        Qwen3-8B \citep{qwen2025qwen25technicalreport} & 36.60 & 31.65 & 78.38 & 48.87 && 51.26 & 42.36 & 77.27 && 40.56 & 34.55 & 81.46 && 39.27 & 35.13 & 76.54 && 15.34 & 14.54 & 78.23\\      
        CodeLlama-70B-Instruct \citep{roziere2023code} & 40.04 & 29.72 & 80.80 & 50.19 && 47.34 & 36.82 & 85.64 && 50.93 & 42.18 & 79.53 && 42.67 & 34.07 & 80.81  && 19.22 & 5.80 & 77.21\\ 
        Deepseek-Chat-V3 \citep{deepseekai2024deepseekv3technicalreport} & 51.47 & 42.26 & \textbf{89.63} & 61.12 && 68.58 & \underline{59.64} & 90.18 && 55.64 & 49.47 & 89.18 && \underline{52.25} & 39.31 & 89.42 && \textbf{29.43} & 20.63 & \textbf{89.72}\\ 
        Deepseek-Coder \citep{guo2024deepseekcoderlargelanguagemodel} & 50.96 & 40.93 & 87.07 & 59.65 && 71.52 & 55.51 & 87.45 && 56.32 & 47.06 & 88.35 && 50.17 & 46.53 & 88.21 &&25.83 & 14.62 & 84.28\\ 
        Deepseek-R1 \citep{deepseekai2025deepseekr1incentivizingreasoningcapability} & \textbf{53.51} & \textbf{45.12} & \underline{89.51} & \textbf{62.71} && \textbf{74.58} & \textbf{60.64} & 90.18 && 57.64 & 48.47 & 89.18 && \textbf{53.39} & \textbf{50.32} & 89.07 && 28.43 & \underline{21.05} & \underline{89.62}\\ 
        Llama3.1-70B \citep{llama3} & 40.33 & 34.40 & 76.52 & 50.42 && 54.05 & 52.36 & 81.82 && 43.56 & 32.71 & 75.76 && 46.12 & 40.32 & 77.25 && 17.57 & 12.20 & 71.23\\ 
        Llama3.1-8B \citep{llama3} & 34.09 & 28.61 & 72.82 & 45.17 && 49.26 & 40.36 & 72.18 && 38.84 & 30.35 & 77.29 && 36.01 & 33.53 & 67.33 && 12.25 & 10.20 & 74.46\\ 
        Llama3.3-70B \citep{llama3} & 42.25 & 31.19 & 78.07 & 50.50 && 56.05 & 49.26 & 82.32 && 46.56 & 31.23 & 78.31 && 48.62 & 31.13 & 78.23 && 18.57 & 13.12 & 73.42\\
        Mistral-Large-Instruct-2407 \citep{jiang2023mistral} & 44.28 & 35.86 & 79.86 & 53.33 && 59.15 & 51.36 & 86.23 && 53.26 & 37.72 & 82.63 && 49.42 & 43.12 & 78.25 && 15.27 & 11.24 & 72.32\\
        Qwen2.5-7B-instruct \citep{qwen2025qwen25technicalreport} & 35.52 & 30.43 & 75.73 & 46.45 && 50.63 & 41.63 & 76.84 && 39.62 & 33.25 & 79.36 && 39.27 & 34.45 & 74.36 && 13.25 & 14.21 & 76.32\\ 
        Telechat2.5-35B \cite{telechat-2024} & 45.18  & 34.71  & 86.56  & 55.48  && 66.45  & 49.98  & 88.32  && 47.12  & 38.12  & 85.23  && 41.02  & 35.07  & 85.84  && 26.13  & 15.67  & 86.85\\
        \midrule
        \multicolumn{3}{l}{\textbf{Closed-Source Models}$^a$} \\
        Moonshot-V1-32K  & 42.41 & 36.05 & 87.11 & 55.19 && 60.25 & 46.55 & 84.36 && 50.35 & 42.24 & 88.35 && 39.72 & 40.20 & 87.20 && 19.33 & 15.21 & 88.54\\ 
        Claude-3.5-Sonnet & 47.62 & 36.31 & 88.61 & 57.51 && 62.18 & 44.36 & 88.43 && 54.28 & 48.53 & 87.59 && 47.64 & 41.13 & 89.60 && 26.39 & 11.24 & 88.83\\ 
        Doubao-Pro-128K & 49.14 & 31.28 & 82.98 & 54.47 && 65.01 & 29.07 & 82.91 && 56.04 & 39.41 & 84.47 && 50.57 & 43.80 & 84.40 && 24.94 & 12.83 & 80.13\\ 
        Doubao-Pro-32K & 44.58 & 31.47 & 81.21 & 52.42 && 61.83 & 34.18 & 79.64 && 51.02 & 44.29 & 82.59 && 48.80 & 37.53 & 83.37 && 16.67 & 9.86 & 79.25\\ 
        GPT-4o \citep{gpt4} & 49.35 & 41.91 & 88.72 & 59.29 && \underline{73.35} & 54.91 & 87.82 && 54.10 & \underline{56.35} & 88.47 && 42.27 & \underline{49.53} & 89.32 && 27.69 & 16.83 & 89.26\\ 
        OpenAI o1-mini & 51.59 & 41.19 & 89.07 & 60.62 && 69.41 & 53.36 & 88.36 && \textbf{60.94} & \textbf{66.29} & 89.53 && 47.98 & 35.27 & \underline{90.17} && 28.04 & 18.84 & 88.21\\ 
        \bottomrule
    \multicolumn{4}{l}{}\\
    \end{tabular}
    }
     
   \caption{Overall performance of LLMs on T2R-bench. For each criterion, the best result is marked in bold, and the second best result is underlined.}
   
   \label{fig:overall result figure}
\end{table*}

\section{Experiments}

\subsection{Experimental Setup}
\textbf{Baselines and Evaluation}.  
We evaluate 25 strong methods on T2R-Bench, including both open-source and closed-source foundation models. The open-source models comprise TableGPT2 \citep{su2024tablegpt2largemultimodalmodel}, Qwen series \citep{qwen, qwen2, qwen2025qwen25technicalreport, hui2024qwen25codertechnicalreport}, Llama family \citep{llama3, roziere2023code}, Mistral \citep{jiang2023mistral}, Deepseek models \citep{deepseekai2024deepseekv3technicalreport,guo2024deepseekcoderlargelanguagemodel, deepseekai2025deepseekr1incentivizingreasoningcapability}, and TeleChat \cite{telechat-2024, sighan-telechat, wang2025technical}, while the closed-source models include GPT series \citep{gpt4}, OpenAI o1-mini, Claude-3.5-Sonnet2, Doubao, and Moonshot. 

The evaluation covers 4 practical industrial scenarios: single tables, multiple tables, complex structured tables, and extremely large-size tables. We assess all models using the proposed metrics: Numerical Accuracy Criterion (NAC), Information Coverage Criterion (ICC), and General Evaluation Criterion (GEC), and we report both overall and average performance scores. \\
\textbf{Implementation Details}. We design a uniform style prompt template to ensure the fairness of the evaluation. Input tables are in Markdown format, and the single LLM directly uses them for generation. For tables whose content exceeds the LLM's context length limit, the content will be truncated. For closed-source models, we utilize official APIs to generate complete reports, with detailed website information provided in Table~\ref{Tab:api} from Appendix~\ref{Appd:websites}. For open-source models, we use 16 A100 40G GPUs for inference. All models use the official default parameters. The uniform style prompt template can be found in Appendix~\ref{Appd:prompt_report}.



\begin{figure}[t]
    \centering
     \includegraphics[width=0.9\linewidth]{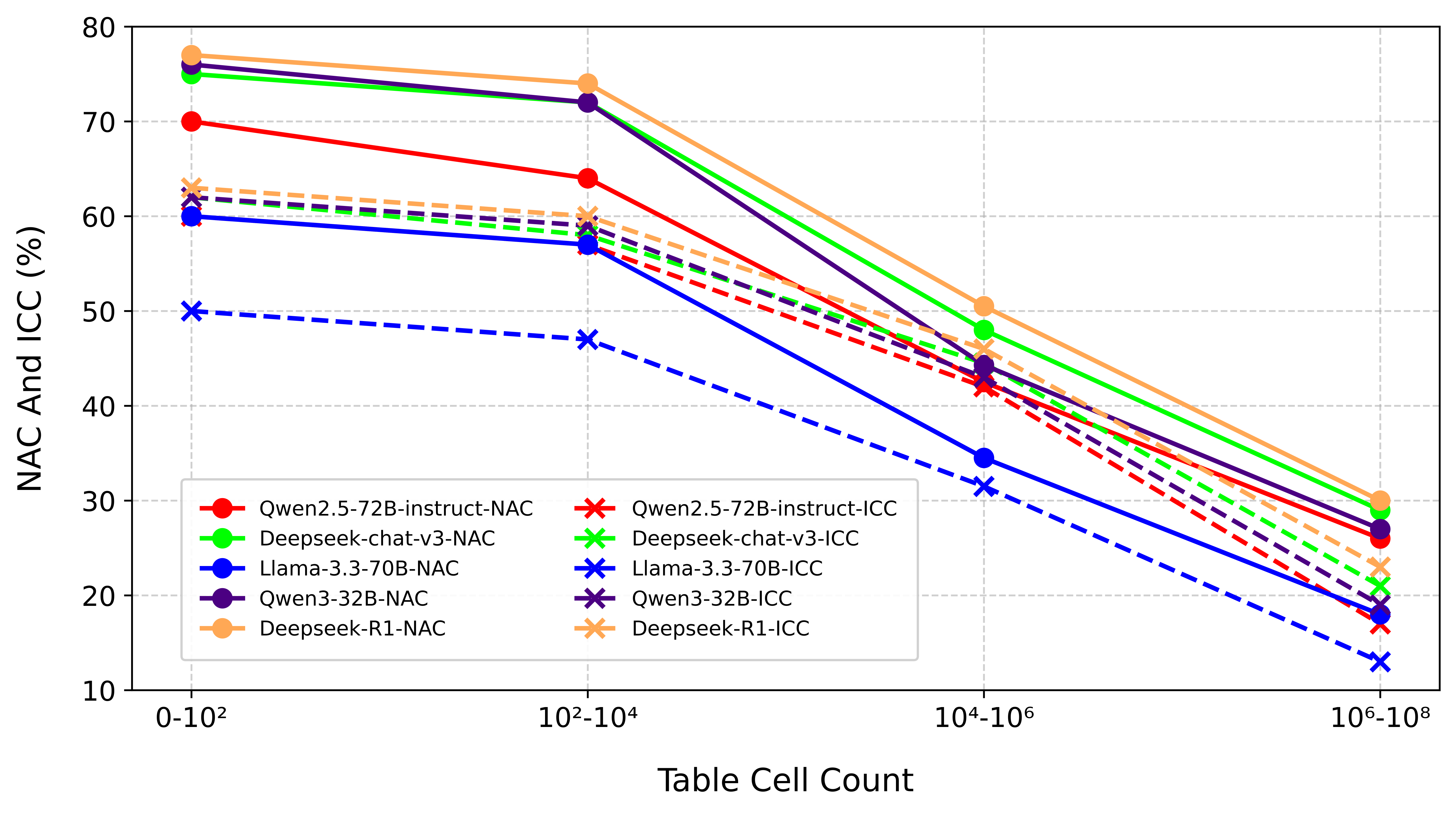}
    \caption{The performance of different LLMs on NAC and ICC criteria across varying numbers of table cell. }
    \label{Fig:IVC_performance_size}
\end{figure}

\begin{table}[t]
    \centering
    \resizebox{0.6\linewidth}{!}{
    \begin{tabular}{l| cc}
        \toprule
        \multirow{2}{*}{\textbf{Model}} & \multicolumn{2}{c}{Languages}\\
         & Chinese & English\\
        \midrule
        Qwen3-32B & 62.43 & 60.07\\ 
        Qwen2.5-72B-Instruct & 60.43 & 58.56\\ 
        Deepseek-R1 & 63.74 & 61.45\\ 
        Llama3.3-70B & 48.26 & 53.24\\ 
        GPT-4o & 59.59 & 60.48\\ 
        \bottomrule
    \end{tabular}}
   \caption{Performance of LLMs on bilingual tables. The indicators in the table are based on the average values of NAC, ICC, and GEC.}
   \label{Tab:bilingual_exp}
\end{table}

\subsection{Main Results}

\textbf{Overall Performance}. As shown in the Table~\ref{fig:overall result figure}, we conduct a comparative analysis of advanced LLMs on the proposed T2R-Bench. We could find: 
(1) The Deepseek series demonstrates superior performance across single table, multiple table, and complex table tasks, establishing its leading capability in Table-to-Report applications. (2) Notably, Qwen3-32B achieves the highest NAC score, showcasing exceptional numerical computation abilities and outperforming even the larger Qwen2.5-72B-Instruct model. (3) While the GPT series maintains strong performance with an ICC score of 66.29\% on multiple table tasks, we observe significant performance degradation across most models when transitioning from single to multiple table tasks, suggesting limitations in cross-table comprehension. (4) The benchmark proves particularly challenging for extremely large-size tables, where all models show substantially reduced performance across all evaluation criteria. The top-performing Deepseek-R1 achieves an average overall score of 62.71\%, highlighting the considerable room for improvement in current approaches for comprehensive table understanding tasks.\\
\textbf{Analysis of Table Cell Count}.
We conduct experiment to investigate how the number of cells in input tables affects the performance. 
As shown in the Figure~\ref{Fig:IVC_performance_size} , we can see that as table size increases, all evaluated LLMs exhibit a sharp performance decline, particularly when processing extremely large-size tables. This finding provides the first empirical evidence in table-related benchmarks that current models face fundamental limitations in comprehending large-scale tabular data, mirroring known challenges in long-text understanding. \\
\textbf{Analysis of Bilingual Capability}.
We conduct the English and Chinese experiment on T2R-Bench, have them processed by the five LLMs for report generation, and subsequently assess using averaged score of the proposed automated evaluation criteria. The Table~\ref{Tab:bilingual_exp} shows that nearly all models exhibit similar performance in both languages, highlighting their consistent ability to generate bilingual reports. However, Llama-3.3-70B's performance in generating Chinese reports lags significantly behind its English capabilities, indicating a need for further fine-tuning. \\
\textbf{Analysis of Input Formatting}. 
Table~\ref{Tab:input_format_exp} demonstrates that among the three most representative table input formats (Markdown, HTML, and JSON), the Markdown format achieves the highest average performance, followed by HTML, while JSON exhibits the lowest performance. 

\begin{table}[t]
    \centering
    \resizebox{0.7\linewidth}{!}{
    \begin{tabular}{cccc}
    \toprule
    ~ & Markdown & Json & Html \\
    \midrule
    Qwen2.5-72B-Instruct & 59.59 & 55.82 & 54.91 \\
    Deepseek-R1 & 62.71 & 58.12 & 60.02 \\ 
    OpenAI-o1-mini & 60.62 & 59.43 & 59.67 \\ 
    \bottomrule
    \end{tabular}}
    \caption{Average performance of NAC, ICC and GEC of three different models across markdown, json and html table input formats.}
    \label{Tab:input_format_exp}
\end{table}

\subsection{Human Evaluation}
\label{Sec: human evaluation}


As table-to-report is a newly formulated task, we establish human baseline for comparison. Given the substantial time commitment required for human report generation, we randomly select a subset of 50 questions (denoted as $D_{val}$) from the dataset by stratified sampling, covering single tables, multiple tables, complex-structure tables, and extremely large-size tables. To mitigate confirmation bias, six independent expert annotators with substantial data analysis experience (and no prior involvement in dataset creation) were recruited to generate reference reports, ensuring unbiased evaluations.


We conducted rigorous validation studies to assess the correlation between our proposed metrics and human evaluation. Another six independent annotators evaluated reports generated by five representative models (Qwen2.5-72B-Instruct, Llama3.3-70B, GPT-4o, DeepSeek-R1, Qwen3-32B-Instruct) alongside human-written reports on $D_{val}$. Evaluations followed criteria (NAC, ICC, GEC from Section~\ref{Sec:criteria}), achieving excellent inter-rater reliability (Fleiss' $k$ = 0.85 \cite{fleiss1973equivalence}). As shown in Table~\ref{table:human}, while systematically more stringent, those metrics demonstrated strong correlation with human judgments (Pearson's $r$ = 0.908 \cite{cohen2009pearson}), validating the framework's reliability despite absolute score differences.
 

\begin{table}[t]
    \centering
    \resizebox{0.9\linewidth}{!}{
    \begin{tabular}{c ccc ccc}
    \toprule 
    Models & Our Evaluation Criteria & Human Evaluation\\  
    \midrule
    Qwen2.5-72B-Instruct & 59.59 & 61.06 \\ 
    Deepseek-R1 & 62.71 & 65.58 \\ 
    Llama3.3-70B & 50.50 & 55.09 \\ 
    GPT-4o & 59.29 & 62.56 \\ 
    Qwen3-32B-Instruct & 61.37 & 63.02\\
    Human baseline & 89.32 & 96.52 \\
    \bottomrule
    \end{tabular}}
    \caption{A consistency test of evaluation methods between the proposed evaluation criteria and human evaluation on average performance of NAC, ICC and GEC.}
    \label{table:human}
\end{table}

\subsection{Case Study}
\begin{figure}[t]
    \centering 
    \includegraphics[width=\linewidth]{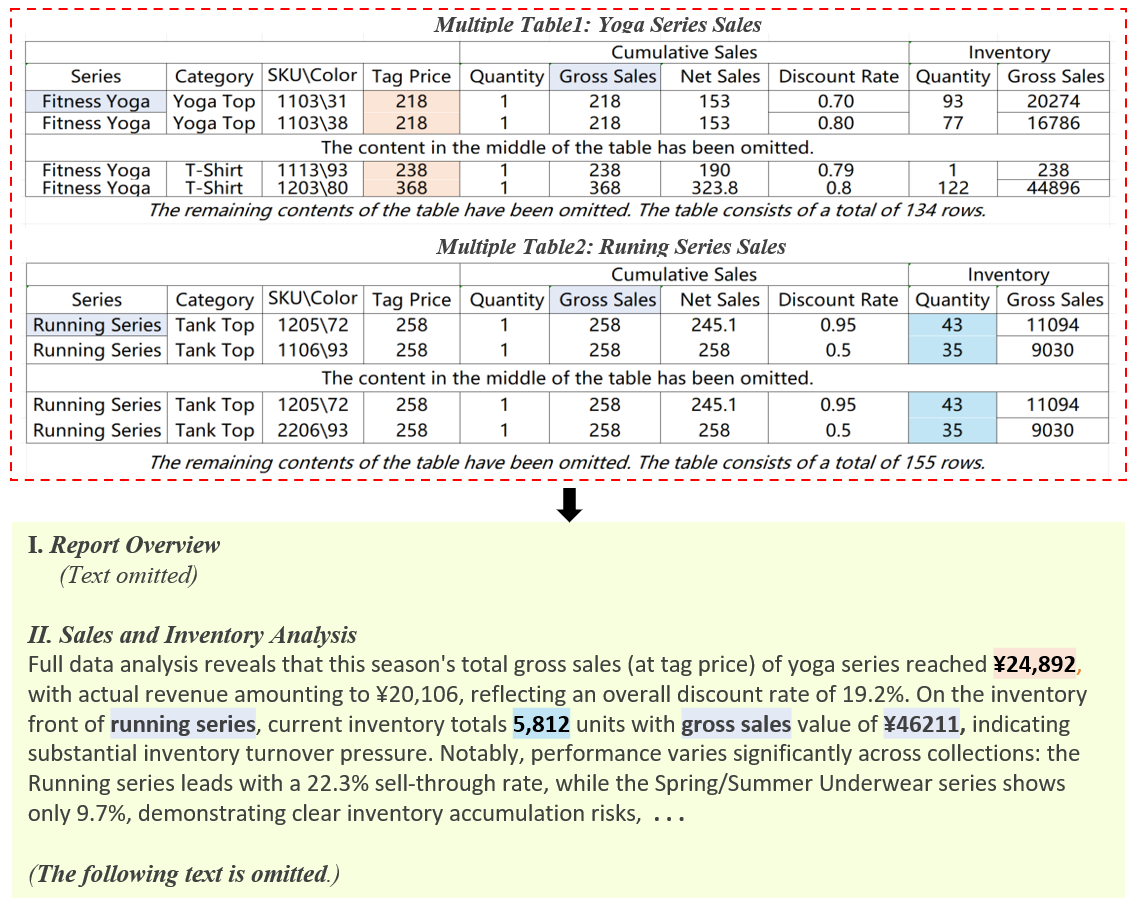} 
    \caption{An example illustrating an original table and its corresponding report generated by DeepSeek-R1, with critical error highlighting.}
    \label{Fig:case_study}
\end{figure}

Our manual analysis of 50 randomly selected error cases from T2R-Bench reveals persistent challenges in LLMs' table-to-report capabilities. 
As shown in Figure~\ref{Fig:case_study}, even the top-performing Deepseek-R1 model exhibits critical failures when processing multiple tables, such as numerical hallucinations (e.g., incorrect summation of "Tag Price" in Table~1) and table selection errors (e.g., mistakenly referencing "Gross Sales" from Table~1 instead of Table~2). These errors, along with challenges posed by complex table structures, descriptive hallucinations, and variable misinterpretations, reveal fundamental reasoning limitations despite the models' ability to generate superficially fluent, human-like reports. Comprehensive case study and error analysis are provided in Appendices \ref{Appd:Case Analysis} and \ref{Appd:Error Analysis}. 

\section{Conclusion}

To meet practical industrial requirements, we introduce the \textbf{table-to-report} task and present \textbf{T2R-bench}, which requires models to generate article-level reports from tabular data. T2R-bench comprises 457 real-world tables spanning 19 diverse domains, with coverage of 4 industrial table types. In addition, we develop an adapted framework to rigorously evaluate model performance and conduct experiments on 25 state-of-the-art LLMs. Experimental results demonstrate that the top-performing model, Deepseek-R1, achieves suboptimal performance, revealing great room for advancing LLMs' capabilities in table-to-report generation.

\section*{Limitations}

While our benchmark represents a significant step forward, several challenges remain. The current best-performing open-source model (Deepseek-R1) achieves suboptimal performance, with both Numerical Accuracy (NAC) and Information Coverage (ICC) scores below 65\% on the proposed evaluation framework. This performance gap highlights two critical needs: (1) the expansion of our benchmark dataset to cover more diverse table types and domains, and (2) the development of specialized models specifically designed for the table-to-report task. These limitations underscore the pressing demand for methodological innovations that can bridge the gap between current capabilities and real-world application requirements.

\section*{Ethics Statement}

In the construction and evaluation of the T2R-Bench, we rigorously adhered to established ethical guidelines for responsible AI research. 
\\
\textbf{Data Collection and Privacy.}
All datasets utilized in this study were sourced from publicly available repositories with potential private and sensitive information eliminated.
\\
\textbf{Annotator Compensation and Instruction.}
Our annotation team comprises 24 annotators, with 12 native English speakers and 12 native Chinese speakers, selected from individuals with extensive experience in analyzing tabular data and demonstrated proficiency in writing analytical reports in relevant fields. We ensure fair compensation for all human annotators, paying each annotator a compensation of \$40 per day, with specialized experts receiving an additional 20\% premium in recognition of their advanced skills. All annotation work is conducted voluntarily with informed consent, and participants were fully aware of the research objectives and data usage policies.

\section*{Acknowledgments}
The work is supported by the National Natural Science Foundation of China (62176029, 62506050), China Postdoctoral Science Foundation Funded Project (2024M763867), Chongqing Higher Education Teaching Reform Research Project (No. 242009).



\bibliography{t2r_emnlp}

\begin{thebibliography}{73}
\providecommand{\natexlab}[1]{#1}

\bibitem[{Bai et~al.(2023)Bai, Bai, Chu, Cui, Dang, Deng, Fan, Ge, Han, Huang, Hui, Ji, Li, Lin, Lin, Liu, Liu, Lu, Lu, Ma, Men, Ren, Ren, Tan, Tan, Tu, Wang, Wang, Wang, Wu, Xu, Xu, Yang, Yang, Yang, Yang, Yao, Yu, Yuan, Yuan, Zhang, Zhang, Zhang, Zhang, Zhou, Zhou, Zhou, and Zhu}]{qwen}
Jinze Bai, Shuai Bai, Yunfei Chu, Zeyu Cui, Kai Dang, Xiaodong Deng, Yang Fan, Wenbin Ge, Yu~Han, Fei Huang, Binyuan Hui, Luo Ji, Mei Li, Junyang Lin, Runji Lin, Dayiheng Liu, Gao Liu, Chengqiang Lu, Keming Lu, and 29 others. 2023.
\newblock \href {https://doi.org/10.48550/ARXIV.2309.16609} {Qwen technical report}.
\newblock \emph{CoRR}, abs/2309.16609.

\bibitem[{Bai et~al.(2024)Bai, Zhang, Lv, Zheng, Zhu, Hou, Dong, Tang, and Li}]{bai2024longwriter}
Yushi Bai, Jiajie Zhang, Xin Lv, Linzhi Zheng, Siqi Zhu, Lei Hou, Yuxiao Dong, Jie Tang, and Juanzi Li. 2024.
\newblock Longwriter: Unleashing 10,000+ word generation from long context llms.
\newblock \emph{arXiv preprint arXiv:2408.07055}.

\bibitem[{Chen(2023)}]{chen2023largelanguagemodelsfew1shot}
Wenhu Chen. 2023.
\newblock \href {https://arxiv.org/abs/2210.06710} {Large language models are few(1)-shot table reasoners}.
\newblock \emph{Preprint}, arXiv:2210.06710.

\bibitem[{Chen et~al.(2019)Chen, Wang, Chen, Zhang, Wang, Li, Zhou, and Wang}]{chen2019tabfact}
Wenhu Chen, Hongmin Wang, Jianshu Chen, Yunkai Zhang, Hong Wang, Shiyang Li, Xiyou Zhou, and William~Yang Wang. 2019.
\newblock Tabfact: A large-scale dataset for table-based fact verification.
\newblock \emph{CoRR}.

\bibitem[{Chen et~al.(2020)Chen, Wang, Chen, Zhang, Wang, Li, Zhou, and Wang}]{chen2020tabfactlargescaledatasettablebased}
Wenhu Chen, Hongmin Wang, Jianshu Chen, Yunkai Zhang, Hong Wang, Shiyang Li, Xiyou Zhou, and William~Yang Wang. 2020.
\newblock \href {https://arxiv.org/abs/1909.02164} {Tabfact: A large-scale dataset for table-based fact verification}.
\newblock \emph{Preprint}, arXiv:1909.02164.

\bibitem[{Cheng et~al.(2022)Cheng, Dong, Wang, Jia, Guo, Gao, Han, Lou, and Zhang}]{cheng2022hitabhierarchicaltabledataset}
Zhoujun Cheng, Haoyu Dong, Zhiruo Wang, Ran Jia, Jiaqi Guo, Yan Gao, Shi Han, Jian-Guang Lou, and Dongmei Zhang. 2022.
\newblock \href {https://arxiv.org/abs/2108.06712} {Hitab: A hierarchical table dataset for question answering and natural language generation}.
\newblock \emph{Preprint}, arXiv:2108.06712.

\bibitem[{Cohen et~al.(2009)Cohen, Huang, Chen, Benesty, Benesty, Chen, Huang, and Cohen}]{cohen2009pearson}
Israel Cohen, Yiteng Huang, Jingdong Chen, Jacob Benesty, Jacob Benesty, Jingdong Chen, Yiteng Huang, and Israel Cohen. 2009.
\newblock Pearson correlation coefficient.
\newblock \emph{Noise reduction in speech processing}, pages 1--4.

\bibitem[{Dai et~al.(2025{\natexlab{a}})Dai, Liu, Zhao, Gao, Sun, and Li}]{dai2025secure}
Muzhi Dai, Shixuan Liu, Zhiyuan Zhao, Junyu Gao, Hao Sun, and Xuelong Li. 2025{\natexlab{a}}.
\newblock Secure tug-of-war (sectow): Iterative defense-attack training with reinforcement learning for multimodal model security.
\newblock \emph{arXiv preprint arXiv:2507.22037}.

\bibitem[{Dai et~al.(2025{\natexlab{b}})Dai, Sun, Zhao, Liu, Li, Gao, and Li}]{dai2025captions}
Muzhi Dai, Jiashuo Sun, Zhiyuan Zhao, Shixuan Liu, Rui Li, Junyu Gao, and Xuelong Li. 2025{\natexlab{b}}.
\newblock From captions to rewards (carevl): Leveraging large language model experts for enhanced reward modeling in large vision-language models.
\newblock \emph{arXiv preprint arXiv:2503.06260}.

\bibitem[{DeepSeek-AI et~al.(2025)DeepSeek-AI, Guo, Yang, Zhang, Song, Zhang, Xu, Zhu, Ma, Wang, Bi, Zhang, Yu, Wu, Wu, Gou, Shao, Li, Gao, Liu, Xue, Wang, Wu, Feng, Lu, Zhao, Deng, Zhang, Ruan, Dai, Chen, Ji, Li, Lin, Dai, Luo, Hao, and et~al}]{deepseekai2025deepseekr1incentivizingreasoningcapability}
DeepSeek-AI, Daya Guo, Dejian Yang, Haowei Zhang, Junxiao Song, Ruoyu Zhang, Runxin Xu, Qihao Zhu, Shirong Ma, Peiyi Wang, Xiao Bi, Xiaokang Zhang, Xingkai Yu, Yu~Wu, Z.~F. Wu, Zhibin Gou, Zhihong Shao, Zhuoshu Li, Ziyi Gao, and 19 others. 2025.
\newblock \href {https://arxiv.org/abs/2501.12948} {Deepseek-r1: Incentivizing reasoning capability in llms via reinforcement learning}.
\newblock \emph{Preprint}, arXiv:2501.12948.

\bibitem[{DeepSeek-AI et~al.(2024)DeepSeek-AI, Liu, Feng, Xue, Wang, Wu, Lu, Zhao, Deng, Zhang, Ruan, Dai, Guo, Yang, Chen, Ji, Li, Lin, Dai, Luo, Hao, Chen, Li, Zhang, Bao, Xu, Wang, Zhang, Ding, Xin, Gao, Li, Qu, Cai, Liang, Guo, Ni, Li, Wang, Chen, Chen, Yuan, Qiu, Li, Song, Dong, Hu, Gao, Guan, Huang, Yu, Wang, Zhang, Xu, Xia, and et~al}]{deepseekai2024deepseekv3technicalreport}
DeepSeek-AI, Aixin Liu, Bei Feng, Bing Xue, Bingxuan Wang, Bochao Wu, Chengda Lu, Chenggang Zhao, Chengqi Deng, Chenyu Zhang, Chong Ruan, Damai Dai, Daya Guo, Dejian Yang, Deli Chen, Dongjie Ji, Erhang Li, Fangyun Lin, Fucong Dai, and 37 others. 2024.
\newblock \href {https://arxiv.org/abs/2412.19437} {Deepseek-v3 technical report}.
\newblock \emph{Preprint}, arXiv:2412.19437.

\bibitem[{Dhingra et~al.(2019)Dhingra, Faruqui, Parikh, Chang, Das, and Cohen}]{dhingra2019handlingdivergentreferencetexts}
Bhuwan Dhingra, Manaal Faruqui, Ankur Parikh, Ming-Wei Chang, Dipanjan Das, and William~W. Cohen. 2019.
\newblock \href {https://arxiv.org/abs/1906.01081} {Handling divergent reference texts when evaluating table-to-text generation}.
\newblock \emph{Preprint}, arXiv:1906.01081.

\bibitem[{Dubey et~al.(2024)Dubey, Jauhri, Pandey, Kadian, Al{-}Dahle, Letman, Mathur, Schelten, Yang, Fan, Goyal, Hartshorn, Yang, Mitra, Sravankumar, Korenev, Hinsvark, Rao, Zhang, Rodriguez, Gregerson, Spataru, Rozi{\`{e}}re, Biron, Tang, Chern, Caucheteux, Nayak, Bi, Marra, McConnell, Keller, Touret, and et~al}]{llama3}
Abhimanyu Dubey, Abhinav Jauhri, Abhinav Pandey, Abhishek Kadian, Ahmad Al{-}Dahle, Aiesha Letman, Akhil Mathur, Alan Schelten, Amy Yang, Angela Fan, Anirudh Goyal, Anthony Hartshorn, Aobo Yang, Archi Mitra, Archie Sravankumar, Artem Korenev, Arthur Hinsvark, Arun Rao, Aston Zhang, and 15 others. 2024.
\newblock \href {https://doi.org/10.48550/ARXIV.2407.21783} {The llama 3 herd of models}.
\newblock \emph{CoRR}, abs/2407.21783.

\bibitem[{Fleiss and Cohen(1973)}]{fleiss1973equivalence}
Joseph~L Fleiss and Jacob Cohen. 1973.
\newblock The equivalence of weighted kappa and the intraclass correlation coefficient as measures of reliability.
\newblock \emph{Educational and psychological measurement}, 33(3):613--619.

\bibitem[{Grijalba et~al.(2024)Grijalba, Lopez, Mart{\'\i}nez-C{\'a}mara, and Camacho-Collados}]{grijalba2024question}
Jorge~Os{\'e}s Grijalba, L~Alfonso~Urena Lopez, Eugenio Mart{\'\i}nez-C{\'a}mara, and Jose Camacho-Collados. 2024.
\newblock Question answering over tabular data with databench: A large-scale empirical evaluation of llms.
\newblock In \emph{Proceedings of the 2024 Joint International Conference on Computational Linguistics, Language Resources and Evaluation (LREC-COLING 2024)}, pages 13471--13488.

\bibitem[{Guo et~al.(2024)Guo, Zhu, Yang, Xie, Dong, Zhang, Chen, Bi, Wu, Li, Luo, Xiong, and Liang}]{guo2024deepseekcoderlargelanguagemodel}
Daya Guo, Qihao Zhu, Dejian Yang, Zhenda Xie, Kai Dong, Wentao Zhang, Guanting Chen, Xiao Bi, Y.~Wu, Y.~K. Li, Fuli Luo, Yingfei Xiong, and Wenfeng Liang. 2024.
\newblock \href {https://arxiv.org/abs/2401.14196} {Deepseek-coder: When the large language model meets programming -- the rise of code intelligence}.
\newblock \emph{Preprint}, arXiv:2401.14196.

\bibitem[{He et~al.(2024)He, Zhou, Xu, Ma, Ding, Du, Gao, Jia, Chen, Han, Yuan, and Zhang}]{Text2Analysis}
Xinyi He, Mengyu Zhou, Xinrun Xu, Xiaojun Ma, Rui Ding, Lun Du, Yan Gao, Ran Jia, Xu~Chen, Shi Han, Zejian Yuan, and Dongmei Zhang. 2024.
\newblock \href {https://doi.org/10.1609/AAAI.V38I16.29779} {Text2analysis: {A} benchmark of table question answering with advanced data analysis and unclear queries}.
\newblock In \emph{Thirty-Eighth {AAAI} Conference on Artificial Intelligence, {AAAI} 2024, Thirty-Sixth Conference on Innovative Applications of Artificial Intelligence, {IAAI} 2024, Fourteenth Symposium on Educational Advances in Artificial Intelligence, {EAAI} 2014, February 20-27, 2024, Vancouver, Canada}, pages 18206--18215. {AAAI} Press.

\bibitem[{Hu et~al.(2024{\natexlab{a}})Hu, Zhao, Wei, Chai, Ma, Wang, Wang, Su, Xu, Zhu, Cheng, Yuan, Li, Kuang, Yang, Yang, and Wu}]{hu2024infiagentdabenchevaluatingagentsdata}
Xueyu Hu, Ziyu Zhao, Shuang Wei, Ziwei Chai, Qianli Ma, Guoyin Wang, Xuwu Wang, Jing Su, Jingjing Xu, Ming Zhu, Yao Cheng, Jianbo Yuan, Jiwei Li, Kun Kuang, Yang Yang, Hongxia Yang, and Fei Wu. 2024{\natexlab{a}}.
\newblock \href {https://arxiv.org/abs/2401.05507} {Infiagent-dabench: Evaluating agents on data analysis tasks}.
\newblock \emph{Preprint}, arXiv:2401.05507.

\bibitem[{Hu et~al.(2024{\natexlab{b}})Hu, Zhao, Wei, Chai, Wang, Wang, Su, Xu, Zhu, Cheng, Yuan, Kuang, Yang, Yang, and Wu}]{InfiAgent-DABench}
Xueyu Hu, Ziyu Zhao, Shuang Wei, Ziwei Chai, Guoyin Wang, Xuwu Wang, Jing Su, Jingjing Xu, Ming Zhu, Yao Cheng, Jianbo Yuan, Kun Kuang, Yang Yang, Hongxia Yang, and Fei Wu. 2024{\natexlab{b}}.
\newblock \href {https://doi.org/10.48550/ARXIV.2401.05507} {Infiagent-dabench: Evaluating agents on data analysis tasks}.
\newblock \emph{CoRR}, abs/2401.05507.

\bibitem[{Hui et~al.(2024)Hui, Yang, Cui, Yang, Liu, Zhang, Liu, Zhang, Yu, Lu, Dang, Fan, Zhang, Yang, Men, Huang, Zheng, Miao, Quan, Feng, Ren, Ren, Zhou, and Lin}]{hui2024qwen25codertechnicalreport}
Binyuan Hui, Jian Yang, Zeyu Cui, Jiaxi Yang, Dayiheng Liu, Lei Zhang, Tianyu Liu, Jiajun Zhang, Bowen Yu, Keming Lu, Kai Dang, Yang Fan, Yichang Zhang, An~Yang, Rui Men, Fei Huang, Bo~Zheng, Yibo Miao, Shanghaoran Quan, and 5 others. 2024.
\newblock \href {https://arxiv.org/abs/2409.12186} {Qwen2.5-coder technical report}.
\newblock \emph{Preprint}, arXiv:2409.12186.

\bibitem[{Jiang et~al.(2023)Jiang, Sablayrolles, Mensch, Bamford, Chaplot, Casas, Bressand, Lengyel, Lample, Saulnier et~al.}]{jiang2023mistral}
Albert~Q Jiang, Alexandre Sablayrolles, Arthur Mensch, Chris Bamford, Devendra~Singh Chaplot, Diego de~las Casas, Florian Bressand, Gianna Lengyel, Guillaume Lample, Lucile Saulnier, and 1 others. 2023.
\newblock Mistral 7b.
\newblock \emph{CoRR}.

\bibitem[{Katsis et~al.(2021)Katsis, Chemmengath, Kumar, Bharadwaj, Canim, Glass, Gliozzo, Pan, Sen, Sankaranarayanan, and Chakrabarti}]{katsis2021aitqaquestionansweringdataset}
Yannis Katsis, Saneem Chemmengath, Vishwajeet Kumar, Samarth Bharadwaj, Mustafa Canim, Michael Glass, Alfio Gliozzo, Feifei Pan, Jaydeep Sen, Karthik Sankaranarayanan, and Soumen Chakrabarti. 2021.
\newblock \href {https://arxiv.org/abs/2106.12944} {Ait-qa: Question answering dataset over complex tables in the airline industry}.
\newblock \emph{Preprint}, arXiv:2106.12944.

\bibitem[{Lebret et~al.(2016)Lebret, Grangier, and Auli}]{lebret2016neuraltextgenerationstructured}
Remi Lebret, David Grangier, and Michael Auli. 2016.
\newblock \href {https://arxiv.org/abs/1603.07771} {Neural text generation from structured data with application to the biography domain}.
\newblock \emph{Preprint}, arXiv:1603.07771.

\bibitem[{Lee et~al.(2024)Lee, Yoon, Jang, Lee, Song, Kim, and Kang}]{DBLP:journals/corr/abs-2410-16848}
Taewhoo Lee, Chanwoong Yoon, Kyochul Jang, Donghyeon Lee, Minju Song, Hyunjae Kim, and Jaewoo Kang. 2024.
\newblock \href {https://doi.org/10.48550/ARXIV.2410.16848} {{ETHIC:} evaluating large language models on long-context tasks with high information coverage}.
\newblock \emph{CoRR}, abs/2410.16848.

\bibitem[{Li et~al.(2024{\natexlab{a}})Li, Jiang, Huang, Beigi, Zhao, Tan, Bhattacharjee, Jiang, Chen, Wu, Shu, Cheng, and Liu}]{li2024llmasajudge}
Dawei Li, Bohan Jiang, Liangjie Huang, Alimohammad Beigi, Chengshuai Zhao, Zhen Tan, Amrita Bhattacharjee, Yuxuan Jiang, Canyu Chen, Tianhao Wu, Kai Shu, Lu~Cheng, and Huan Liu. 2024{\natexlab{a}}.
\newblock From generation to judgment: Opportunities and challenges of llm-as-a-judge.
\newblock \emph{arXiv preprint arXiv: 2411.16594}.

\bibitem[{Li et~al.(2024{\natexlab{b}})Li, Du, Zheng, and Song}]{li2024mimotable}
Zheng Li, Yang Du, Mao Zheng, and Mingyang Song. 2024{\natexlab{b}}.
\newblock \href {https://arxiv.org/abs/2412.11711} {Mimotable: A multi-scale spreadsheet benchmark with meta operations for table reasoning}.

\bibitem[{Li et~al.(2024{\natexlab{c}})Li, Wu, Li, He, Fang, Zhang, Zhao, Li, Li, and Song}]{li2024scalable}
Zhongqiu Li, Zhenhe Wu, Mengxiang Li, Zhongjiang He, Ruiyu Fang, Jie Zhang, Yu~Zhao, Yongxiang Li, Zhoujun Li, and Shuangyong Song. 2024{\natexlab{c}}.
\newblock Scalable database-driven kgs can help text-to-sql.
\newblock In \emph{Proceedings of the {ISWC} 2024 Posters, Demos and Industry Tracks: From Novel Ideas to Industrial Practice co-located with 23nd International Semantic Web Conference {(ISWC} 2024), Hanover, Maryland, USA, November 11-15, 2024}, volume 3828 of \emph{{CEUR} Workshop Proceedings}. CEUR-WS.org.

\bibitem[{Lin(2004)}]{lin2004rouge}
Chin-Yew Lin. 2004.
\newblock \href {https://aclanthology.org/W04-1013/} {{ROUGE}: A package for automatic evaluation of summaries}.
\newblock In \emph{Text Summarization Branches Out}, pages 74--81, Barcelona, Spain. Association for Computational Linguistics.

\bibitem[{Lu et~al.(2024)Lu, Zhang, Zhang, and Chen}]{Survey}
Weizheng Lu, Jiaming Zhang, Jing Zhang, and Yueguo Chen. 2024.
\newblock \href {https://doi.org/10.48550/ARXIV.2402.05121} {Large language model for table processing: {A} survey}.
\newblock \emph{CoRR}, abs/2402.05121.

\bibitem[{Ma et~al.(2023)Ma, Ding, Wang, Han, and Zhang}]{ma-etal-2023-insightpilot}
Pingchuan Ma, Rui Ding, Shuai Wang, Shi Han, and Dongmei Zhang. 2023.
\newblock \href {https://doi.org/10.18653/v1/2023.emnlp-demo.31} {{I}nsight{P}ilot: An {LLM}-empowered automated data exploration system}.
\newblock In \emph{Proceedings of the 2023 Conference on Empirical Methods in Natural Language Processing: System Demonstrations}, pages 346--352.

\bibitem[{Ma et~al.(2024)Ma, Zhang, Zhang, Yu, Zhang, Zhang, Luo, Wang, and Tang}]{ma2024spreadsheetbenchchallengingrealworld}
Zeyao Ma, Bohan Zhang, Jing Zhang, Jifan Yu, Xiaokang Zhang, Xiaohan Zhang, Sijia Luo, Xi~Wang, and Jie Tang. 2024.
\newblock \href {https://arxiv.org/abs/2406.14991} {Spreadsheetbench: Towards challenging real world spreadsheet manipulation}.
\newblock \emph{Preprint}, arXiv:2406.14991.

\bibitem[{Mathur et~al.(2024)Mathur, Siu, Lipka, and Sun}]{mathur-etal-2024-matsa}
Puneet Mathur, Alexa Siu, Nedim Lipka, and Tong Sun. 2024.
\newblock \href {https://doi.org/10.18653/v1/2024.emnlp-demo.26} {{MATSA}: Multi-agent table structure attribution}.
\newblock In \emph{Proceedings of the 2024 Conference on Empirical Methods in Natural Language Processing: System Demonstrations}, pages 250--258, Miami, Florida, USA. Association for Computational Linguistics.

\bibitem[{Nan et~al.(2021)Nan, Hsieh, Mao, Lin, Verma, Zhang, Kryściński, Schoelkopf, Kong, Tang, Mutuma, Rosand, Trindade, Bandaru, Cunningham, Xiong, and Radev}]{nan2021fetaqafreeformtablequestion}
Linyong Nan, Chiachun Hsieh, Ziming Mao, Xi~Victoria Lin, Neha Verma, Rui Zhang, Wojciech Kryściński, Nick Schoelkopf, Riley Kong, Xiangru Tang, Murori Mutuma, Ben Rosand, Isabel Trindade, Renusree Bandaru, Jacob Cunningham, Caiming Xiong, and Dragomir Radev. 2021.
\newblock \href {https://arxiv.org/abs/2104.00369} {Fetaqa: Free-form table question answering}.
\newblock \emph{Preprint}, arXiv:2104.00369.

\bibitem[{OpenAI(2023)}]{gpt4}
OpenAI. 2023.
\newblock \href {https://arxiv.org/abs/2303.08774} {Gpt-4 technical report}.
\newblock \emph{arXiv preprint arXiv:2303.08774}.

\bibitem[{Os{\'e}s~Grijalba et~al.(2024)Os{\'e}s~Grijalba, Ure{\~n}a-L{\'o}pez, Mart{\'i}nez~C{\'a}mara, and Camacho-Collados}]{oses-grijalba-etal-2024-question}
Jorge Os{\'e}s~Grijalba, L.~Alfonso Ure{\~n}a-L{\'o}pez, Eugenio Mart{\'i}nez~C{\'a}mara, and Jose Camacho-Collados. 2024.
\newblock \href {https://aclanthology.org/2024.lrec-main.1179/} {Question answering over tabular data with {D}ata{B}ench: A large-scale empirical evaluation of {LLM}s}.
\newblock In \emph{Proceedings of the 2024 Joint International Conference on Computational Linguistics, Language Resources and Evaluation (LREC-COLING 2024)}, pages 13471--13488, Torino, Italia. ELRA and ICCL.

\bibitem[{Papineni et~al.(2002{\natexlab{a}})Papineni, Roukos, Ward, and Zhu}]{Papineni2002BleuAM}
Kishore Papineni, Salim Roukos, Todd Ward, and Wei-Jing Zhu. 2002{\natexlab{a}}.
\newblock \href {https://api.semanticscholar.org/CorpusID:11080756} {Bleu: a method for automatic evaluation of machine translation}.
\newblock In \emph{Annual Meeting of the Association for Computational Linguistics}.

\bibitem[{Papineni et~al.(2002{\natexlab{b}})Papineni, Roukos, Ward, and Zhu}]{papineni2002bleu}
Kishore Papineni, Salim Roukos, Todd Ward, and Wei-Jing Zhu. 2002{\natexlab{b}}.
\newblock Bleu: a method for automatic evaluation of machine translation.
\newblock In \emph{Proceedings of the 40th annual meeting of the Association for Computational Linguistics}, pages 311--318.

\bibitem[{Parikh et~al.(2020)Parikh, Wang, Gehrmann, Faruqui, Dhingra, Yang, and Das}]{ToTTo}
Ankur~P. Parikh, Xuezhi Wang, Sebastian Gehrmann, Manaal Faruqui, Bhuwan Dhingra, Diyi Yang, and Dipanjan Das. 2020.
\newblock \href {https://doi.org/10.18653/V1/2020.EMNLP-MAIN.89} {Totto: {A} controlled table-to-text generation dataset}.
\newblock In \emph{Proceedings of the 2020 Conference on Empirical Methods in Natural Language Processing, {EMNLP} 2020, Online, November 16-20, 2020}, pages 1173--1186. Association for Computational Linguistics.

\bibitem[{Pasupat and Liang(2015{\natexlab{a}})}]{WikiTableQuestion}
Panupong Pasupat and Percy Liang. 2015{\natexlab{a}}.
\newblock \href {https://doi.org/10.3115/V1/P15-1142} {Compositional semantic parsing on semi-structured tables}.
\newblock In \emph{Proceedings of the 53rd Annual Meeting of the Association for Computational Linguistics and the 7th International Joint Conference on Natural Language Processing of the Asian Federation of Natural Language Processing, {ACL} 2015, July 26-31, 2015, Beijing, China, Volume 1: Long Papers}, pages 1470--1480. The Association for Computer Linguistics.

\bibitem[{Pasupat and Liang(2015{\natexlab{b}})}]{pasupat-liang-2015-compositional}
Panupong Pasupat and Percy Liang. 2015{\natexlab{b}}.
\newblock \href {https://doi.org/10.3115/v1/P15-1142} {Compositional semantic parsing on semi-structured tables}.
\newblock In \emph{Proceedings of the 53rd Annual Meeting of the Association for Computational Linguistics and the 7th International Joint Conference on Natural Language Processing (Volume 1: Long Papers)}, pages 1470--1480, Beijing, China. Association for Computational Linguistics.

\bibitem[{Qwen et~al.(2025)Qwen, :, Yang, Yang, Zhang, Hui, Zheng, Yu, Li, Liu, Huang, Wei, Lin, Yang, Tu, Zhang, Yang, Yang, Zhou, Lin, Dang, Lu, Bao, Yang, Yu, Li, Xue, Zhang, Zhu, Men, Lin, Li, Tang, Xia, Ren, Ren, Fan, Su, Zhang, Wan, Liu, Cui, Zhang, and Qiu}]{qwen2025qwen25technicalreport}
Qwen, :, An~Yang, Baosong Yang, Beichen Zhang, Binyuan Hui, Bo~Zheng, Bowen Yu, Chengyuan Li, Dayiheng Liu, Fei Huang, Haoran Wei, Huan Lin, Jian Yang, Jianhong Tu, Jianwei Zhang, Jianxin Yang, Jiaxi Yang, Jingren Zhou, and 25 others. 2025.
\newblock \href {https://arxiv.org/abs/2412.15115} {Qwen2.5 technical report}.
\newblock \emph{Preprint}, arXiv:2412.15115.

\bibitem[{Roziere et~al.(2023)Roziere, Gehring, Gloeckle, Sootla, Gat, Tan, Adi, Liu, Remez, Rapin et~al.}]{roziere2023code}
Baptiste Roziere, Jonas Gehring, Fabian Gloeckle, Sten Sootla, Itai Gat, Xiaoqing~Ellen Tan, Yossi Adi, Jingyu Liu, Tal Remez, J{\'e}r{\'e}my Rapin, and 1 others. 2023.
\newblock Code llama: Open foundation models for code.
\newblock \emph{CORR}.

\bibitem[{Shao and Li(2025)}]{aiflow-2024}
Jiawei Shao and Xuelong Li. 2025.
\newblock \href {https://doi.org/10.1109/MNET.2025.3541208} {Ai flow at the network edge}.
\newblock \emph{IEEE Network}.

\bibitem[{Su et~al.(2024)Su, Wang, Ye, Zhou, Zhang, Chen, Zhu, Wang, Xu, Chen, Li, Lan, Tian, Yuan, Zhao, Zhou, Shou, Zha, Long, Li, Wu, Zhang, Huang, Yang, Zhang, Ye, Zhu, Hu, Gu, Sun, Li, Yang, and Xiao}]{su2024tablegpt2largemultimodalmodel}
Aofeng Su, Aowen Wang, Chao Ye, Chen Zhou, Ga~Zhang, Gang Chen, Guangcheng Zhu, Haobo Wang, Haokai Xu, Hao Chen, Haoze Li, Haoxuan Lan, Jiaming Tian, Jing Yuan, Junbo Zhao, Junlin Zhou, Kaizhe Shou, Liangyu Zha, Lin Long, and 14 others. 2024.
\newblock \href {https://arxiv.org/abs/2411.02059} {Tablegpt2: A large multimodal model with tabular data integration}.
\newblock \emph{Preprint}, arXiv:2411.02059.

\bibitem[{Sui et~al.(2024)Sui, Zhou, Zhou, Han, and Zhang}]{SUC}
Yuan Sui, Mengyu Zhou, Mingjie Zhou, Shi Han, and Dongmei Zhang. 2024.
\newblock \href {https://doi.org/10.1145/3616855.3635752} {Table meets {LLM:} can large language models understand structured table data? {A} benchmark and empirical study}.
\newblock In \emph{Proceedings of the 17th {ACM} International Conference on Web Search and Data Mining, {WSDM} 2024, Merida, Mexico, March 4-8, 2024}, pages 645--654. {ACM}.

\bibitem[{Szymanski et~al.(2025)Szymanski, Ziems, Eicher{-}Miller, Li, Jiang, and Metoyer}]{DBLP:conf/iui/SzymanskiZEL0M25}
Annalisa Szymanski, Noah Ziems, Heather~A. Eicher{-}Miller, Toby~Jia{-}Jun Li, Meng Jiang, and Ronald~A. Metoyer. 2025.
\newblock \href {https://doi.org/10.1145/3708359.3712091} {Limitations of the llm-as-a-judge approach for evaluating {LLM} outputs in expert knowledge tasks}.
\newblock In \emph{Proceedings of the 30th International Conference on Intelligent User Interfaces, {IUI} 2025, Cagliari, Italy, March 24-27, 2025}, pages 952--966. {ACM}.

\bibitem[{Tang et~al.(2024)Tang, Zong, Phang, Zhao, Zhou, Cohan, and Gerstein}]{tang-etal-2024-struc}
Xiangru Tang, Yiming Zong, Jason Phang, Yilun Zhao, Wangchunshu Zhou, Arman Cohan, and Mark Gerstein. 2024.
\newblock \href {https://doi.org/10.18653/v1/2024.naacl-short.2} {Struc-bench: Are large language models good at generating complex structured tabular data?}
\newblock In \emph{Proceedings of the 2024 Conference of the North American Chapter of the Association for Computational Linguistics: Human Language Technologies (Volume 2: Short Papers)}, pages 12--34, Mexico City, Mexico. Association for Computational Linguistics.

\bibitem[{Wang et~al.(2024{\natexlab{a}})Wang, Liu, Zheng, Qi, Chen, Yang, Zhao, Wang, Song, and Huang}]{wang-etal-2024-boosting-llm}
Shenzhi Wang, Chang Liu, Zilong Zheng, Siyuan Qi, Shuo Chen, Qisen Yang, Andrew Zhao, Chaofei Wang, Shiji Song, and Gao Huang. 2024{\natexlab{a}}.
\newblock \href {https://doi.org/10.18653/v1/2024.findings-acl.591} {Boosting {LLM} agents with recursive contemplation for effective deception handling}.
\newblock In \emph{Findings of the Association for Computational Linguistics: ACL 2024}, pages 9909--9953, Bangkok, Thailand. Association for Computational Linguistics.

\bibitem[{Wang et~al.(2025{\natexlab{a}})Wang, Yu, Gao, Zheng, Liu, Lu, Dang, Chen, Yang, Zhang et~al.}]{wang2025beyond}
Shenzhi Wang, Le~Yu, Chang Gao, Chujie Zheng, Shixuan Liu, Rui Lu, Kai Dang, Xionghui Chen, Jianxin Yang, Zhenru Zhang, and 1 others. 2025{\natexlab{a}}.
\newblock Beyond the 80/20 rule: High-entropy minority tokens drive effective reinforcement learning for llm reasoning.
\newblock \emph{arXiv preprint arXiv:2506.01939}.

\bibitem[{Wang et~al.()Wang, Fang, Li, He, and Song}]{wang2025when}
Shiquan Wang, Ruiyu Fang, Mengxiang Li, Zhongjiang He, and Shuangyong Song.
\newblock When less is more: Minimal prompts with lora for llm text detection.
\newblock In \emph{The 14th CCF International Conference on Natural Language Processing and Chinese Computing}.

\bibitem[{Wang et~al.(2023)Wang, Kordi, Mishra, Liu, Smith, Khashabi, and Hajishirzi}]{wang-etal-2023-self-instruct}
Yizhong Wang, Yeganeh Kordi, Swaroop Mishra, Alisa Liu, Noah~A. Smith, Daniel Khashabi, and Hannaneh Hajishirzi. 2023.
\newblock \href {https://doi.org/10.18653/v1/2023.acl-long.754} {Self-instruct: Aligning language models with self-generated instructions}.
\newblock In \emph{Proceedings of the 61st Annual Meeting of the Association for Computational Linguistics (Volume 1: Long Papers)}, pages 13484--13508, Toronto, Canada. Association for Computational Linguistics.

\bibitem[{Wang et~al.(2024{\natexlab{b}})Wang, Liu, Liu, Yao, Huang, He, Li, Li, Che, Zhang, Wang, Wang, Pu, Xu, Fang, Zhao, Zhang, Huang, Lu, Peng, Zheng, Wang, Yang, He, Jiang, Xie, Zhang, Li, Shi, Fu, Zhang, Huang, Xiong, Zhang, Wang, and Song}]{telechat-2024}
Zihan Wang, Xinzhang Liu, Shixuan Liu, Yitong Yao, Yuyao Huang, Zhongjiang He, Xuelong Li, Yongxiang Li, Zhonghao Che, Zhaoxi Zhang, Yan Wang, Xin Wang, Luwen Pu, Huihan Xu, Ruiyu Fang, Yu~Zhao, Jie Zhang, Xiaomeng Huang, Zhilong Lu, and 17 others. 2024{\natexlab{b}}.
\newblock \href {https://arxiv.org/abs/2401.03804} {Telechat technical report}.
\newblock \emph{{Computing Research Repository}}.

\bibitem[{Wang et~al.(2024{\natexlab{c}})Wang, Liu, Liu, Yao, Huang, Li, He, Li, Pu, Xu, Wang, and Song}]{sighan-telechat}
Zihan Wang, XinZhang Liu, Shixuan Liu, Yitong Yao, Yuyao Huang, Mengxiang Li, Zhongjiang He, Yongxiang Li, Luwen Pu, Huinan Xu, Chao Wang, and Shuangyong Song. 2024{\natexlab{c}}.
\newblock Telechat: An open-source billingual large language model.
\newblock In \emph{Proceedings of the 10th SIGHAN Workshop on Chinese Language Processing (SIGHAN-10)}.

\bibitem[{Wang et~al.(2025{\natexlab{b}})Wang, Liu, Yao, Wang, Zhao, Yang, Deng, Jia, Peng, Huang et~al.}]{wang2025technical}
Zihan Wang, Xinzhang Liu, Yitong Yao, Chao Wang, Yu~Zhao, Zhihao Yang, Wenmin Deng, Kaipeng Jia, Jiaxin Peng, Yuyao Huang, and 1 others. 2025{\natexlab{b}}.
\newblock Technical report of telechat2, telechat2. 5 and t1.
\newblock \emph{arXiv preprint arXiv:2507.18013}.

\bibitem[{Wei et~al.(2023{\natexlab{a}})Wei, Sun, Zhang, Jin, Zhang, Lv, and Guo}]{DBLP:journals/tkde/WeiSZJZLG23}
Kaiwen Wei, Xian Sun, Zequn Zhang, Li~Jin, Jingyuan Zhang, Jianwei Lv, and Zhi Guo. 2023{\natexlab{a}}.
\newblock \href {https://doi.org/10.1109/TKDE.2022.3218830} {Implicit event argument extraction with argument-argument relational knowledge}.
\newblock \emph{{IEEE} Trans. Knowl. Data Eng.}, 35(9):8865--8879.

\bibitem[{Wei et~al.(2021)Wei, Sun, Zhang, Zhang, Guo, and Jin}]{DBLP:conf/acl/WeiSZZGJ20}
Kaiwen Wei, Xian Sun, Zequn Zhang, Jingyuan Zhang, Zhi Guo, and Li~Jin. 2021.
\newblock \href {https://doi.org/10.18653/V1/2021.ACL-LONG.360} {Trigger is not sufficient: Exploiting frame-aware knowledge for implicit event argument extraction}.
\newblock In \emph{Proceedings of the 59th Annual Meeting of the Association for Computational Linguistics and the 11th International Joint Conference on Natural Language Processing, {ACL/IJCNLP} 2021, (Volume 1: Long Papers), Virtual Event, August 1-6, 2021}, pages 4672--4682. Association for Computational Linguistics.

\bibitem[{Wei et~al.(2023{\natexlab{b}})Wei, Yang, Jin, Sun, Zhang, Zhang, Li, Zhang, Liu, and Guo}]{DBLP:conf/acl/WeiYJSZZLZLZ23}
Kaiwen Wei, Yiran Yang, Li~Jin, Xian Sun, Zequn Zhang, Jingyuan Zhang, Xiao Li, Linhao Zhang, Jintao Liu, and Zhi Guo. 2023{\natexlab{b}}.
\newblock \href {https://doi.org/10.18653/V1/2023.ACL-LONG.272} {Guide the many-to-one assignment: Open information extraction via iou-aware optimal transport}.
\newblock In \emph{Proceedings of the 61st Annual Meeting of the Association for Computational Linguistics (Volume 1: Long Papers), {ACL} 2023, Toronto, Canada, July 9-14, 2023}, pages 4971--4984. Association for Computational Linguistics.

\bibitem[{Wiseman et~al.(2017)Wiseman, Shieber, and Rush}]{wiseman2017challengesdatatodocumentgeneration}
Sam Wiseman, Stuart~M. Shieber, and Alexander~M. Rush. 2017.
\newblock \href {https://arxiv.org/abs/1707.08052} {Challenges in data-to-document generation}.
\newblock \emph{Preprint}, arXiv:1707.08052.

\bibitem[{Wu et~al.(2024)Wu, Yang, Chai, Zhang, Liu, Du, Liang, Shu, Cheng, Sun et~al.}]{wu2024tablebench}
Xianjie Wu, Jian Yang, Linzheng Chai, Ge~Zhang, Jiaheng Liu, Xinrun Du, Di~Liang, Daixin Shu, Xianfu Cheng, Tianzhen Sun, and 1 others. 2024.
\newblock Tablebench: A comprehensive and complex benchmark for table question answering.
\newblock \emph{arXiv preprint arXiv:2408.09174}.

\bibitem[{Wu et~al.(2025{\natexlab{a}})Wu, Li, Li, Zhang, He, Yang, Zhao, Fang, Li, Li, and Song}]{MR-SQL}
Zhenhe Wu, Zhongqiu Li, Mengxiang Li, Jie Zhang, Zhongjiang He, Jian Yang, Yu~Zhao, Ruiyu Fang, Yongxiang Li, Zhoujun Li, and Shuangyong Song. 2025{\natexlab{a}}.
\newblock {MR-SQL:} multi-level retrieval enhances inference for llm in text-to-sql.
\newblock \emph{DASFAA}.

\bibitem[{Wu et~al.(2025{\natexlab{b}})Wu, Li, Zhang, He, Yang, Zhao, Fang, Wang, Xie, Song, and Li}]{wu2025uniting}
Zhenhe Wu, Zhongqiu Li, Jie Zhang, Zhongjiang He, Jian Yang, Yu~Zhao, Ruiyu Fang, Bing Wang, Hongyan Xie, Shuangyong Song, and Zhoujun Li. 2025{\natexlab{b}}.
\newblock {UCS-SQL:} uniting content and structure for enhanced semantic bridging in text-to-sql.
\newblock In \emph{Findings of the Association for Computational Linguistics, {ACL} 2025, Vienna, Austria, July 27 - August 1, 2025}, pages 8156--8168. Association for Computational Linguistics.

\bibitem[{Wu et~al.(2025{\natexlab{c}})Wu, Yang, Liu, Wu, Pan, Zhang, Zhao, Song, Li, and Li}]{wu2025table}
Zhenhe Wu, Jian Yang, Jiaheng Liu, Xianjie Wu, Changzai Pan, Jie Zhang, Yu~Zhao, Shuangyong Song, Yongxiang Li, and Zhoujun Li. 2025{\natexlab{c}}.
\newblock Table-r1: Region-based reinforcement learning for table understanding.
\newblock \emph{arXiv preprint arXiv:2505.12415}.

\bibitem[{Xing et~al.(2025)Xing, Liu, Jiang, Yang, Yao, Wang, Deng, Wang, Song, Yang et~al.}]{xing2025llmsr}
Hongrui Xing, Xinzhang Liu, Zhuo Jiang, Zhihao Yang, Yitong Yao, Zihan Wang, Wenmin Deng, Chao Wang, Shuangyong Song, Wang Yang, and 1 others. 2025.
\newblock Llmsr@ xllm25: A language model-based pipeline for structured reasoning data construction.
\newblock In \emph{Proceedings of the 1st Joint Workshop on Large Language Models and Structure Modeling (XLLM 2025)}, pages 342--350.

\bibitem[{Xiong et~al.(2025{\natexlab{a}})Xiong, Li, Wang, Zhao, Zhang, Pan, He, Li, Chang, He et~al.}]{xiong2025teleai}
Sishi Xiong, Mengxiang Li, Dakai Wang, Yu~Zhao, Jie Zhang, Changzai Pan, Haowei He, Xiangyu Li, Wenhan Chang, Zhongjiang He, and 1 others. 2025{\natexlab{a}}.
\newblock Teleai at semeval-2025 task 8: Advancing table reasoning framework with large language models.
\newblock In \emph{Proceedings of the 19th International Workshop on Semantic Evaluation (SemEval-2025)}, pages 1828--1841.

\bibitem[{Xiong et~al.(2025{\natexlab{b}})Xiong, Wang, Zhao, Zhang, Pan, He, Li, Chang, He, Song et~al.}]{xiong2025tablereasoner}
Sishi Xiong, Dakai Wang, Yu~Zhao, Jie Zhang, Changzai Pan, Haowei He, Xiangyu Li, Wenhan Chang, Zhongjiang He, Shuangyong Song, and 1 others. 2025{\natexlab{b}}.
\newblock Tablereasoner: Advancing table reasoning framework with large language models.
\newblock \emph{arXiv preprint arXiv:2507.08046}.

\bibitem[{Yang et~al.(2024{\natexlab{a}})Yang, Yang, Hui, Zheng, Yu, Zhou, Li, Li, Liu, Huang, Dong, Wei, Lin, Tang, Wang, Yang, Tu, Zhang, Ma, Yang, Xu, Zhou, Bai, He, Lin, Dang, Lu, Chen, Yang, Li, Xue, Ni, Zhang, Wang, Peng, Men, Gao, Lin, Wang, Bai, Tan, Zhu, Li, Liu, Ge, Deng, Zhou, Ren, Zhang, Wei, Ren, Liu, Fan, Yao, Zhang, Wan, Chu, Liu, Cui, Zhang, Guo, and Fan}]{qwen2}
An~Yang, Baosong Yang, Binyuan Hui, Bo~Zheng, Bowen Yu, Chang Zhou, Chengpeng Li, Chengyuan Li, Dayiheng Liu, Fei Huang, Guanting Dong, Haoran Wei, Huan Lin, Jialong Tang, Jialin Wang, Jian Yang, Jianhong Tu, Jianwei Zhang, Jianxin Ma, and 43 others. 2024{\natexlab{a}}.
\newblock \href {https://doi.org/10.48550/ARXIV.2407.10671} {Qwen2 technical report}.
\newblock \emph{CoRR}, abs/2407.10671.

\bibitem[{Yang et~al.(2024{\natexlab{b}})Yang, Liu, and Kan}]{yang-etal-2024-datatales}
Yajing Yang, Qian Liu, and Min-Yen Kan. 2024{\natexlab{b}}.
\newblock \href {https://doi.org/10.18653/v1/2024.emnlp-main.601} {{D}ata{T}ales: A benchmark for real-world intelligent data narration}.
\newblock In \emph{Proceedings of the 2024 Conference on Empirical Methods in Natural Language Processing}, pages 10764--10788, Miami, Florida, USA. Association for Computational Linguistics.

\bibitem[{Yu et~al.(2018)Yu, Zhang, Yang, Yasunaga, Wang, Li, Ma, Li, Yao, Roman, Zhang, and Radev}]{yu-etal-2018-spider}
Tao Yu, Rui Zhang, Kai Yang, Michihiro Yasunaga, Dongxu Wang, Zifan Li, James Ma, Irene Li, Qingning Yao, Shanelle Roman, Zilin Zhang, and Dragomir Radev. 2018.
\newblock \href {https://doi.org/10.18653/v1/D18-1425} {{S}pider: A large-scale human-labeled dataset for complex and cross-domain semantic parsing and text-to-{SQL} task}.
\newblock In \emph{Proceedings of the 2018 Conference on Empirical Methods in Natural Language Processing}, pages 3911--3921, Brussels, Belgium. Association for Computational Linguistics.

\bibitem[{Zhang et~al.(2020)Zhang, Kishore, Wu, Weinberger, and Artzi}]{zhang2020bertscoreevaluatingtextgeneration}
Tianyi Zhang, Varsha Kishore, Felix Wu, Kilian~Q. Weinberger, and Yoav Artzi. 2020.
\newblock \href {https://arxiv.org/abs/1904.09675} {Bertscore: Evaluating text generation with bert}.
\newblock \emph{Preprint}, arXiv:1904.09675.

\bibitem[{Zhao et~al.(2025)Zhao, Han, Wu, He, Ning, Yuan, Li, Wang, and Song}]{zhao2025enhancing}
Deji Zhao, Donghong Han, Jia Wu, Zhongjiang He, Bo~Ning, Ye~Yuan, Yongxiang Li, Chao Wang, and Shuangyong Song. 2025.
\newblock Enhancing math reasoning ability of large language models via computation logic graphs.
\newblock \emph{Knowledge-Based Systems}, page 113905.

\bibitem[{Zheng et~al.(2023)Zheng, Chiang, Sheng, Zhuang, Wu, Zhuang, Lin, Li, Li, Xing et~al.}]{zheng2023judging}
Lianmin Zheng, Wei-Lin Chiang, Ying Sheng, Siyuan Zhuang, Zhanghao Wu, Yonghao Zhuang, Zi~Lin, Zhuohan Li, Dacheng Li, Eric Xing, and 1 others. 2023.
\newblock Judging llm-as-a-judge with mt-bench and chatbot arena.
\newblock \emph{Advances in Neural Information Processing Systems}, 36:46595--46623.

\bibitem[{Zhong et~al.(2017)Zhong, Xiong, and Socher}]{zhong2017seq2sqlgeneratingstructuredqueries}
Victor Zhong, Caiming Xiong, and Richard Socher. 2017.
\newblock \href {https://arxiv.org/abs/1709.00103} {Seq2sql: Generating structured queries from natural language using reinforcement learning}.
\newblock \emph{Preprint}, arXiv:1709.00103.

\bibitem[{Zhu et~al.(2021)Zhu, Lei, Huang, Wang, Zhang, Lv, Feng, and Chua}]{zhu-etal-2021-tat}
Fengbin Zhu, Wenqiang Lei, Youcheng Huang, Chao Wang, Shuo Zhang, Jiancheng Lv, Fuli Feng, and Tat-Seng Chua. 2021.
\newblock \href {https://doi.org/10.18653/v1/2021.acl-long.254} {{TAT}-{QA}: A question answering benchmark on a hybrid of tabular and textual content in finance}.
\newblock In \emph{Proceedings of the 59th Annual Meeting of the Association for Computational Linguistics and the 11th International Joint Conference on Natural Language Processing (Volume 1: Long Papers)}, pages 3277--3287, Online. Association for Computational Linguistics.

\end{thebibliography}

\newpage
\appendix

\section{Examples of T2R-bench}

\subsection{English Table Example with Report Generated by the Single LLM}
This subsection shows an example of a $<$question, table, report keypoints$>$ triple, with report generated through the single LLM method of Qwen2.5-72B-Instruct. The incorrect parts in the report have been highlighted in red.

\begin{tcolorbox}[sidebyside, sidebyside align=top seam, width=\linewidth, colback=gray!20, colframe=white, colbacktitle=white, coltitle=white, breakable, arc=0mm, left=0mm, right=0mm]
\small
\textbf{Question}\\

Formulate an insightful report entitled ``Handler Effect Evaluation'', examining the efficiency of various handlers in managing stock entries and departures as recorded in January 2020's data.\\

\textbf{Table}\\

\begin{minipage}{\linewidth}
    \centering 
    \resizebox{\linewidth}{!}{
    \begin{tabular}{|l|l|l|l|c|c|c|c|l|l|}
    \hline
        \thead{Serial\\Number} & \thead{Date} & \thead{Product\\Name} & \thead{ID} & \multicolumn{2}{c|}{\thead{Stock\\In}} & \multicolumn{2}{c|}{\thead{Stock\\Out}} & \thead{Real-time\\Inventory} & \thead{Handler} \\
        \cline{5-8}
         &  &  &  & Quantity & Supplier & Quantity & Supplier &  &  \\
    \hline
        1 & 1/10/20 & Name1 & mg-022 & 16 & Supplier01 & 8 & Supplier01 & 8 & Wang Jing \\
    \hline
        2 & 1/11/20 & Name2 & mg-023 & 62 & Supplier02 & 51 & Supplier02 & 11 & Wang Jing \\
    \hline
        3 & 1/12/20 & Name3 & mg-024 & 23 & Supplier03 & 15 & Supplier03 & 8 & Wang Jing \\
    \hline
        4 & 1/13/20 & Name4 & mg-025 & 126 & Supplier04 & 85 & Supplier04 & 41 & Li Kaihua \\
    \hline
        5 & 1/14/20 & Name5 & mg-026 & 103 & Supplier05 & 55 & Supplier05 & 48 & Li Kaihua \\
    \hline
        6 & 1/15/20 & Name6 & mg-027 & 85 & Supplier06 & 27 & Supplier06 & 58 & Li Kaihua \\
    \hline
        7 & 1/16/20 & Name7 & mg-028 & 54 & Supplier07 & 38 & Supplier07 & 16 & Li Kaihua \\
    \hline
    \end{tabular}}
    \captionof{table}{Inventory Management Table} 
\end{minipage}
\\\\

\textbf{Report Reference Keypoints}\\

Keypoint 1: Wang Jing's average incoming quantity is 33.67, and the average outgoing quantity is 24.67, indicating a gradual increase in inventory, but the final inventory level remains stable.\\

Keypoint 2: Li Kaihua's average incoming quantity is 92.0, and the average outgoing quantity is 51.25. Despite the significantly higher incoming quantity compared to the outgoing quantity, the actual inventory level has decreased in practice.\\

Keypoint 3: Wang Jing's inventory management is more volatile, with an average real-time inventory of 9.0, indicating a lower and relatively stable inventory level.\\

Keypoint 4: Li Kaihua's inventory activities are more active, with an average real-time inventory of 40.75, indicating a higher inventory level.\\

Keypoint 5: Wang Jing's real-time inventory data for the earliest and latest dates processed is 8, indicating that the inventory level remained stable during the processing period with no significant changes.\\

Keypoint 6: Li Kaihua's initial inventory was 41.0, and the final inventory decreased to 16.0, indicating a reduction in actual inventory levels, possibly due to unrecorded outgoing activities or other factors.\\

Keypoint 7: Wang Jing's inventory management is relatively stable, but the inventory level is low, indicating a more conservative management strategy.\\

Keypoint 8: Li Kaihua's inventory management strategy is more active, with a higher average incoming quantity, but the actual inventory level has decreased, indicating a potentially more aggressive management strategy.\\

\textbf{Generated Report}\\

Title: Handler Effect Evaluation\\

Macro Description

This report aims to evaluate the efficiency of different handlers in managing inventory in and out during the period from January 10, 2020 to January 16, 2020. The data covers 7 inventory records involving 7 different products, and each record includes information such as date, product name, product ID, inbound quantity, outbound quantity, real-time inventory, and handler. By analyzing these data, we can understand the performance of different handlers in inventory management, thereby providing a basis for optimizing the inventory management process.\\

Overall Situation of Inbound and Outbound

During the period from January 10, 2020 to January 16, 2020, a total of 7 inventory operations were recorded, involving 7 different products. Overall, in these 7 operations, the total inbound quantity was \textcolor{red}{460} pieces, the total outbound quantity was \textcolor{red}{274} pieces, and the final real-time inventory was 186 pieces. Specifically, the inbound quantity on January 10 was 16 pieces, and the outbound quantity was 8 pieces; the inbound quantity on January 11 was 62 pieces, and the outbound quantity was 51 pieces; the inbound quantity on January 12 was 23 pieces, and the outbound quantity was 15 pieces; the inbound quantity on January 13 was 126 pieces, and the outbound quantity was 85 pieces; the inbound quantity on January 14 was 103 pieces, and the outbound quantity was 55 pieces; the inbound quantity on January 15 was 85 pieces, and the outbound quantity was 27 pieces; the inbound quantity on January 16 was 54 pieces, and the outbound quantity was 38 pieces. These data indicate that, overall, the inbound quantity is slightly higher than the outbound quantity, and the inventory has increased.\\

Handler Efficiency Analysis

In the 7 inventory operations, the handlers are mainly divided into two: Wang Jing and Li Kaihua. Wang Jing handled 3 operations, and Li Kaihua handled 4 operations. Specifically, in the 3 operations handled by Wang Jing, the total inbound quantity was \textcolor{red}{91} pieces, the total outbound quantity was 74 pieces, and the final real-time inventory was \textcolor{red}{17} pieces. In the 4 operations handled by Li Kaihua, the total inbound quantity was \textcolor{red}{369} pieces, the total outbound quantity was \textcolor{red}{200} pieces, and the final real-time inventory was \textcolor{red}{169} pieces. From these data, it can be seen that the inbound and outbound quantities handled by Li Kaihua are much higher than those handled by Wang Jing, and the final real-time inventory is also significantly higher than that of Wang Jing. This indicates that Li Kaihua is more efficient in handling a large number of inventory operations.\\

Specific Performance of Handlers

Further analysis of the specific performance of Wang Jing and Li Kaihua reveals some interesting phenomena. In the 3 operations handled by Wang Jing, the inbound and outbound quantities of each operation are relatively small, and the real-time inventory changes after each operation are not significant. For example, the inbound quantity on January 10 was 16 pieces, the outbound quantity was 8 pieces, and the real-time inventory was 8 pieces; the inbound quantity on January 11 was 62 pieces, the outbound quantity was 51 pieces, and the real-time inventory was 11 pieces; the inbound quantity on January 12 was 23 pieces, the outbound quantity was 15 pieces, and the real-time inventory was 8 pieces. This indicates that Wang Jing performs stably when handling small-batch inventory operations, but has relatively weak ability to handle large-batch inventory operations.\\

In contrast, in the 4 operations handled by Li Kaihua, the inbound and outbound quantities of each operation are relatively large, and the real-time inventory changes after each operation are obvious. For example, the inbound quantity on January 13 was 126 pieces, the outbound quantity was 85 pieces, and the real-time inventory was 41 pieces; the inbound quantity on January 14 was 103 pieces, the outbound quantity was 55 pieces, and the real-time inventory was 48 pieces; the inbound quantity on January 15 was 85 pieces, the outbound quantity was 27 pieces, and the real-time inventory was 58 pieces; the inbound quantity on January 16 was 54 pieces, the outbound quantity was 38 pieces, and the real-time inventory was 16 pieces. This indicates that Li Kaihua is more efficient in handling large-batch inventory operations and can complete tasks quickly and accurately.\\

Comparison of Handler Efficiency

In order to more intuitively compare the efficiency of Wang Jing and Li Kaihua, we can calculate the processing time of each operation. Assuming that the processing time of each operation is the same, the average inbound quantity of Wang Jing in handling 3 operations is \textcolor{red}{30.33} pieces, and the average outbound quantity is 24.67 pieces; the average inbound quantity of Li Kaihua in handling 4 operations is 92.25 pieces, and the average outbound quantity is \textcolor{red}{50} pieces. From these data, it can be seen that the average inbound and outbound quantities of Li Kaihua in each operation are significantly higher than those of Wang Jing, which further confirms the efficiency of Li Kaihua in handling large-batch inventory operations.\\

In addition, we can also calculate the real-time inventory change after each operation. In the 3 operations handled by Wang Jing, the real-time inventory changes after each operation are \textcolor{red}{8 pieces, -41 pieces, and -7 pieces}, respectively, with an average change of \textcolor{red}{-13.33} pieces; in the 4 operations handled by Li Kaihua, the real-time inventory changes after each operation are \textcolor{red}{41 pieces, -7 pieces, 31 pieces, and -22 pieces}, respectively, with an average change of \textcolor{red}{10.75} pieces. This indicates that Li Kaihua can better maintain the stability of inventory after each operation, while Wang Jing, although performing stably when handling small-batch inventory operations, may cause greater inventory fluctuations when handling large-batch inventory operations.\\

Summary and Recommendations

In summary, through the analysis of inventory management data from January 10, 2020 to January 16, 2020, we can draw the following conclusions: Li Kaihua is more efficient in handling large-batch inventory operations, can complete tasks quickly and accurately, and can better maintain the stability of inventory after each operation. In contrast, Wang Jing performs stably when handling small-batch inventory operations, but may cause greater inventory fluctuations when handling large-batch inventory operations.\\

Based on the above analysis, we make the following recommendations:\\
1. **Optimize personnel allocation**: When handling large-batch inventory operations, prioritize Li Kaihua for operation to improve overall efficiency and inventory stability. For small-batch inventory operations, Wang Jing can continue to be responsible to maintain the stability and accuracy of the operation.\\
2. **Training and improvement**: In view of the shortcomings of Wang Jing in handling large-batch inventory operations, special training can be arranged to improve her ability to handle large-batch inventory operations. At the same time, regular performance evaluations can also be used to motivate handlers to continuously improve their skills.\\
3. **Process optimization**: In the inventory management process, more automated tools and technologies can be introduced to reduce the error of manual operations and improve overall efficiency. For example, an inventory management system can be used to automatically record and update inventory data to reduce the burden of manual operations.\\

Through the above measures, the inventory management process can be further optimized, the overall efficiency and inventory stability can be improved, and better support can be provided for the operation of the enterprise.
\end{tcolorbox}

\subsection{Chinese Table Example with Report Generated by the Single LLM}

This subsection shows an example of a $<$question, table, report keypoints$>$ triple, with report generated through the single LLM of Qwen2.5-72B-Instruct. The incorrect parts in the report have been highlighted in red.

\begin{tcolorbox}[sidebyside, sidebyside align=top seam, width=\linewidth, colback=gray!20, colframe=white, colbacktitle=white, coltitle=white, breakable, arc=0mm, left=0mm, right=0mm]
\small
\textbf{Question}\\

\begin{CJK}{UTF8}{gbsn}
统计不同地区农村家庭卫生设施差异，形成《农村卫生设备区域对比》，重点探讨水冲式厕所在各区域的普及程度。\\
\end{CJK}

\textbf{Table}\\

\begin{CJK}{UTF8}{gbsn}
\begin{minipage}{\linewidth}
    \centering
    \resizebox{\linewidth}{!}{
    \begin{tabular}{|l|l|l|l|l|l|}
    \hline
        ~ & 饮用水情况 & 饮用水情况 & 饮用水情况 & 饮用水情况 & 饮用水情况  \\
        ~ & 单位：\% & 单位：\% & 单位：\% & 单位：\% & 单位：\%  \\
        ~ & 全国 & 东部地区 & 中部地区 & 西部地区 & 东北地区  \\ \hline
        获取饮用水困难住户的比重 & 10.3 & 2.8 & 9.6 & 22.2 & 1.3  \\ \hline
        使用管道水住户的比重 & 48.6 & 71.1 & 28.5 & 42.9 & 41.3  \\ \hline
        净化处理过的饮用水 & 23.1 & 44.2 & 8.8 & 13.3 & 15.0  \\ \hline
        深井水 & 41.8 & 37.6 & 53.4 & 27.5 & 75.9  \\ \hline
        浅井水 & 27.8 & 15.3 & 31.7 & 43.6 & 9.0  \\ \hline
        江河湖水 & 2.8 & 1.4 & 2.0 & 6.0 & 0.1  \\ \hline
        池塘水 & 1.4 & 0.4 & 1.6 & 2.6 & 0.0  \\ \hline
        雨水 & 1.4 & 0.1 & 1.1 & 3.7 & 0.0  \\ \hline
        其他水 & 1.7 & 1.0 & 1.4 & 3.3 & 0.0  \\ \hline
    \end{tabular}}
    \captionof{table}{农村生活条件与设备普及率统计。}
\end{minipage}
\end{CJK}\\\\

\textbf{Report Reference Keypoints}\\

\begin{CJK}{UTF8}{gbsn}
关键点1：全国农村家庭中，仅有12.8\%的家庭拥有水冲式厕所，旱厕和简易厕所或无厕所的家庭占比87.2\%。\\
关键点2：东部地区水冲式厕所普及率最高，为26.0\%，而东北地区最低，仅为1.3\%。\\
关键点3：中部地区旱厕最为普遍，占54.1\%。\\
关键点4：西部地区农村家庭中，简易厕所或无厕所的比例最高，达到53.2\%，水冲式厕所比例仅为6.2\%。\\
\end{CJK}

\textbf{Generated Report}\\

\begin{CJK}{UTF8}{gbsn}
《农村卫生设备区域对比》\\

宏观描述\\
本报告基于全国及各地区农村家庭卫生设施的数据，重点探讨了水冲式厕所在不同区域的普及程度。数据显示，全国范围内水冲式厕所的普及率仅为12.8\%，而不同地区的普及率差异显著。东部地区的普及率最高，达到26.0\%，而东北地区的普及率最低，仅为1.3\%。这一差异反映了各地区在卫生设施建设和管理上的不同水平，也揭示了农村卫生设施改善的迫切需求。\\

东部地区：水冲式厕所普及率最高\\
东部地区的水冲式厕所普及率达到了26.0\%，远高于全国平均水平。这一高普及率的背后，是东部地区相对发达的经济条件和较高的居民生活水平。东部地区的农村家庭不仅在卫生设施上投入更多，而且在其他生活设施方面也表现出较高的普及率。例如，东部地区的彩电普及率为97.5台/百户，固定电话普及率为68.2部/百户，手机普及率为86.1部/百户，电脑普及率为4.8台/百户，摩托车普及率为50.9辆/百户，生活用汽车普及率为5.1辆/百户。这些数据表明，东部地区的农村家庭在基础设施和生活条件方面整体较好，为水冲式厕所的普及提供了良好的基础。\\

此外，东部地区的饮用水条件也较为优越。使用管道水的住户比例为71.1\%，远高于全国平均水平的48.6\%。净化处理过的饮用水的使用比例为44.2\%，同样高于全国平均水平的23.1\%。这些数据进一步说明，东部地区的农村家庭在卫生和生活条件方面具有明显优势，为水冲式厕所的普及提供了有力支持。\\

中部地区：水冲式厕所普及率较低\\
中部地区的水冲式厕所普及率仅为6.0\%，远低于全国平均水平。这一低普及率反映了中部地区在卫生设施建设和管理上的不足。尽管中部地区的经济条件相对较好，但卫生设施的普及率仍然较低。中部地区的彩电普及率为85.1台/百户，固定电话普及率为45.9部/百户，手机普及率为68.7部/百户，电脑普及率为0.9台/百户，摩托车普及率为36辆/百户，生活用汽车普及率为2.6辆/百户。这些数据表明，中部地区的农村家庭在基础设施和生活条件方面虽然有所改善，但仍存在较大差距。\\

中部地区的饮用水条件也较为一般。使用管道水的住户比例为28.5\%，低于全国平均水平的48.6\%。净化处理过的饮用水的使用比例为8.8\%，远低于全国平均水平的23.1\%。这些数据说明，中部地区的农村家庭在卫生和生活条件方面仍需进一步提升，特别是在水冲式厕所的普及方面。\\

西部地区：水冲式厕所普及率略高于中部\\
西部地区的水冲式厕所普及率为6.2\%，略高于中部地区的6.0\%，但仍低于全国平均水平。西部地区的经济条件相对较为落后，基础设施建设相对滞后。西部地区的彩电普及率为74.5台/百户，固定电话普及率为35.1部/百户，手机普及率为52.3部/百户，电脑普及率为0.6台/百户，摩托车普及率为25.6辆/百户，生活用汽车普及率为2.2辆/百户。这些数据表明，西部地区的农村家庭在基础设施和生活条件方面存在较大差距，特别是在卫生设施方面。\\

西部地区的饮用水条件也较为落后。使用管道水的住户比例为42.9\%，低于全国平均水平的48.6\%。净化处理过的饮用水的使用比例为13.3\%，略高于中部地区的8.8\%，但仍低于全国平均水平的23.1\%。这些数据说明，西部地区的农村家庭在卫生和生活条件方面仍需进一步改善，特别是在水冲式厕所的普及方面。\\

东北地区：水冲式厕所普及率最低\\
东北地区的水冲式厕所普及率仅为1.3\%，是全国最低的。这一低普及率反映了东北地区在卫生设施建设和管理上的严重不足。尽管东北地区的经济条件相对较好，但卫生设施的普及率仍然较低。东北地区的彩电普及率为97.1台/百户，固定电话普及率为64.4部/百户，手机普及率为63.7部/百户，电脑普及率为1台/百户，摩托车普及率为34.3辆/百户，生活用汽车普及率为2.6辆/百户。这些数据表明，东北地区的农村家庭在基础设施和生活条件方面虽然有所改善，但仍存在较大差距。\\

东北地区的饮用水条件也较为一般。使用管道水的住户比例为41.3\%，略低于全国平均水平的48.6\%。\textcolor{red}{净化处理过的饮用水的使用比例为15.0\%，略高于全国平均水平的23.1\%。}这些数据说明，东北地区的农村家庭在卫生和生活条件方面仍需进一步提升，特别是在水冲式厕所的普及方面。
\end{CJK}
\end{tcolorbox}

\section{Implementation Details for Benchmark Construction}

\subsection{Details for Annotation Team Composition}
\label{Appd:annotation}

We recruit a total of 24 annotators in three batches, evenly split between native Chinese and English speakers. All annotators hold Master’s degrees and have at least one year of experience in data analysis and report writing.

The first group of 12 annotators focuses on dataset construction and annotation introduced in Section~\ref{Sec:bench construction}, including six senior specialists with domain-specific report writing experience across six distinct fields in the dataset. These senior members serve as quality control reviewers, conducting final verification of annotations to ensure accuracy and consistency throughout the dataset development process.

The second group comprises six evaluators responsible for human evaluation of generated reports introduced in Section~\ref{Sec: human evaluation}. This team receive comprehensive training through virtual meetings to establish unified evaluation criteria, enabling them to systematically annotate and score reports based on predefined quality metrics while maintaining inter-rater reliability.

The third group contains six independent report writers who manually create reference reports serving as the human baseline introduced in Section~\ref{Sec: human evaluation}. This isolated team operates without exposure to the dataset construction details or evaluation protocols, ensuring an objective performance baseline by preventing any potential information leakage that might influence their writing outputs.

All annotators work eight hours a day and earned a wage of \$40 per day on average, with specialized experts receiving an additional 20\% premium. All annotators are trained through videos or online meetings provided with annotation guidelines that explains the data usage for academic research purposes.

\subsection{Details of Procedure for Question Annotation}
\label{Appd:annotation_procedure}
We randomly assign each question to two annotators, whose selection criteria and qualifications are detailed in Section~\ref{Appd:annotation}.

Each annotator assesses the quality of question candidate based on the following aspects: \textbf{a) scope compliance}: the question must be answerable using tabular data, without requiring any extraneous domain knowledge. Temporal and spatial references must be strictly confined within the boundaries of the dataset. \textbf{b) thematic focus}: the question should concentrate on a single analytical dimension to derive evidence-bound conclusions, rather than enabling the generation of multi-thematic reports across divergent analytical directions. \textbf{c) conceptual distinctiveness}: multiple questions derived from the same table must address non-overlapping thematic aspects with clearly differentiated analytical objectives.

In cases where the evaluation results of the two annotators are inconsistent, the results will be handed over to a third annotator for the final judgment. Through this rigorous quality assurance procedure, we obtained 910 high-quality, comprehensive questions.

\subsection{Details of Procedure for Keypoints Annotation}
\label{Appd:key_points_annotation}

Similarly to question annotation, each <table, question> pair and corresponding three groups of extracted report key points is assigned to two independent annotators for revision. 
However, more complicated than binary validity judgments in question annotation, key point annotation requires multi-dimensional modifications including summarization, deletion, insertion and polishing based on AI-generated report keypoints. The annotation of key points adheres to the following criteria: 1) Factual Accuracy: The keypoints must be derived from and accurately reflect the data presented in the tables. 2) Relevance: The keypoints must align with the question of the report generation. 3) Essentiality: The key points should encompass the core content necessary to address the report's objectives.
4) Consistency: The key points should be logically coherent, non-repetitive, and form a cohesive narrative.

The results of two annotators are assigned to the third annotator for justification. If the third annotator finds the two annotations to be consistent or very similar, they will make minor adjustments and approve it as the final core point. However, if the third annotator identifies significant discrepancies between the two annotations, the issue will be documented and discussed during the daily meeting to reach a consensus with the other two annotators.

\subsection{Prompts Library and Seed Questions for Question Generation}
\label{Appd:question_gernation_prompt}

The five prompt templates in the prompt library for question generation are shown below:

\begin{tcolorbox}[sidebyside, sidebyside align=top seam, width=\linewidth, colback=gray!20, colframe=white, colbacktitle=white, coltitle=white, breakable, arc=0mm, left=0mm, right=0mm]
\small
As an expert with extensive experience in data analysis and report writing, you are required to propose questions for generating reports from multiple different perspectives along with specific requirements, based on the table description uploaded. The questions must be detailed enough and ensure there is sufficient differentiation among the questions.\\

\#\# Response Format:\\

\hspace{15pt}
\begin{minipage}{\dimexpr\linewidth-15pt}
Question 1:...\\
Question 2:...\\
Question 3:...\\
\end{minipage}

\#\# Input Table Descriptions:\\
{\red [TABLE DESCRIPTION]}\\

Please directly output the generated 3 questions, do not include any additional explanations or comments.
\end{tcolorbox}

\begin{tcolorbox}[sidebyside, sidebyside align=top seam, width=\linewidth, colback=gray!20, colframe=white, colbacktitle=white, coltitle=white, breakable, arc=0mm, left=0mm, right=0mm]
\small
As an expert in table structure comprehension and narrative synthesis, develop three questions that cover different investigative angles, such as ratio and share analysis, comparative growth rates, and anomaly flagging—detailing the fields involved, the analytical approach.\\

\#\# Response Format:\\

\hspace{15pt}
\begin{minipage}{\dimexpr\linewidth-15pt}
Question 1:...\\
Question 2:...\\
Question 3:...\\
\end{minipage}

\#\# Input Table Descriptions:\\
{\red [TABLE DESCRIPTION]}\\

Please directly output the generated 3 questions, do not include any additional explanations or comments.
\end{tcolorbox}

\begin{tcolorbox}[sidebyside, sidebyside align=top seam, width=\linewidth, colback=gray!20, colframe=white, colbacktitle=white, coltitle=white, breakable, arc=0mm, left=0mm, right=0mm]
\small
Drawing on your ability to unpick multidimensional tables and deliver actionable insights, craft three report questions that each emphasize a unique focus—layered segmentation, extreme-value exploration, and temporal dynamics—while clarifying which metrics to calculate, over what time or dimension range, and the expected outcome for managerial decision support.\\

\#\# Response Format:\\

\hspace{15pt}
\begin{minipage}{\dimexpr\linewidth-15pt}
Question 1:...\\
Question 2:...\\
Question 3:...\\
\end{minipage}

\#\# Input Table Descriptions:\\
{\red [TABLE DESCRIPTION]}\\

Please directly output the generated 3 questions, do not include any additional explanations or comments.
\end{tcolorbox}

\begin{tcolorbox}[sidebyside, sidebyside align=top seam, width=\linewidth, colback=gray!20, colframe=white, colbacktitle=white, coltitle=white, breakable, arc=0mm, left=0mm, right=0mm]
\small
As a specialist skilled in dissecting complex tables and translating data into clear narratives, design three probing questions that focus respectively on trend analysis, distributional characteristics, and comparative benchmarking; for each, indicate the key metric, the comparison group or baseline, the required data granularity, and the decision-making context.\\

\#\# Response Format:\\

\hspace{15pt}
\begin{minipage}{\dimexpr\linewidth-15pt}
Question 1:...\\
Question 2:...\\
Question 3:...\\
\end{minipage}

\#\# Input Table Descriptions:\\
{\red [TABLE DESCRIPTION]}\\

Please directly output the generated 3 questions, do not include any additional explanations or comments.
\end{tcolorbox}

\begin{tcolorbox}[sidebyside, sidebyside align=top seam, width=\linewidth, colback=gray!20, colframe=white, colbacktitle=white, coltitle=white, breakable, arc=0mm, left=0mm, right=0mm]
\small
From the perspective of an analyst with deep expertise in interpreting tabular datasets and crafting concise reports, propose three questions that each target a different analytical dimension, such as time series trends, category comparisons, or geographic breakdowns—while specifying the exact fields to use, the calculations or aggregations required.\\

\#\# Response Format:\\

\hspace{15pt}
\begin{minipage}{\dimexpr\linewidth-15pt}
Question 1:...\\
Question 2:...\\
Question 3:...\\
\end{minipage}

\#\# Input Table Descriptions:\\
{\red [TABLE DESCRIPTION]}\\

Please directly output the generated 3 questions, do not include any additional explanations or comments.
\end{tcolorbox}

\noindent The 10 Seed Questions are shown below:

\begin{tcolorbox}[sidebyside, sidebyside align=top seam, width=\linewidth, colback=gray!20, colframe=white, colbacktitle=white, coltitle=white, breakable, arc=0mm, left=0mm, right=0mm]
\small

Seed Question 1:Produce a report entitled 'Analysis of Stock Market Trading Trends in May 2006', providing a comprehensive examination of monthly fluctuations in both trading volume and transaction amounts.\\

Seed Question 2:Develop 'Q3 2008 Metals and Fuel Oil Market Dynamics Report', investigating annual trading value fluctuations and market trends for copper, aluminum, and zinc contracts on SHFE.\\

Seed Question 3:Analyze September 2023 food and alcohol price variations across China's major cities, with particular focus on how grain and vegetable price movements impact composite indices.\\

Seed Question 4:Prepare an in-depth report on 'Model Differentiation Analysis for Shenbei Avenue 4S Stores (July)', evaluating sales performance and customer preference across vehicle models.\\

Seed Question 5:Generate a trend analysis report on township hospital bed utilization rates from 2014 to 2022, utilizing comprehensive tabular data.\\

Seed Question 6:Conduct 'Historical Analysis of Healthcare Personnel Structure (2014-2023)', tracking growth patterns across medical staff categories with special attention to licensed physicians and registered nurses.\\

Seed Question 7:Complete 'Human Resource Efficiency in Small and Micro Enterprises', examining workforce allocation and revenue efficiency across industries using employment and income data.\\

Seed Question 8:Produce 'Study on Melon Cultivation Structure Transformation (2014-2022)', detailing area changes for watermelon, muskmelon, strawberry and related crops.\\

Seed Question 9:Investigate productivity trends in petroleum and natural gas extraction, creating a detailed change analysis report for August 2023 through March 2024.\\

Seed Question 10:Compile a report titled 'Seasonal Fluctuation Analysis of Beijing Secondary Housing Prices', conducting year-round data dissection with emphasis on seasonal influencing factors.
\end{tcolorbox}

\subsection{Prompt for Report Keypoints Extraction}
\label{Appd:report_core_points_extraction}
\begin{tcolorbox}[sidebyside, sidebyside align=top seam, width=\linewidth, colback=gray!20, colframe=white, colbacktitle=white, coltitle=white, breakable, arc=0mm, left=0mm, right=0mm]
\small
As an expert with extensive experience in information extraction, you are required to summarize 5-10 report reference keypoints based on the report I provide. \\

\#\# Response Format:\\
\hspace{15pt}
\begin{minipage}{\dimexpr\linewidth-15pt}
Keypoint 1:...\\
Keypoint 2:...\\
\end{minipage}

\#\# Reference Reports:\\
{\red [REPORTS]}\\

Please directly output the generated 5-10 report reference keypoints, do not include any additional explanations or comments.
\end{tcolorbox}

\subsection{Domain and Sub-domain of T2R-bench}
\label{Appd:domain_distribution}
The 6 domains and 19 sub-domains in T2R-bench are shown in Table~\ref{Tab:domains}.

\begin{table}[ht]
   \centering
    \resizebox{\linewidth}{!}{
    \begin{tabular}{ll}
    \toprule
    \textbf{Domains} & \textbf{Sub-domains} \\ 
    \midrule
    \makecell[l]{Engineering \\ Science} & \makecell[l]{Electronics and Automation Manufacturing;\\ Chemical Engineering and Advanced Materials;\\ Energy Production and Power Systems;\\ Automotive Manufacturing and Mobility Solutions}  \\ 
    \midrule
    \makecell[l]{Environmental \\Stewardship} & \makecell[l]{Environmental Protection;\\Agricultural Production and Forestry Management; \\Marine Resources and Fisheries Management}  \\ 
    \midrule
    \makecell[l]{Transportation\\Logistics} & \makecell[l]{Telecommunications and IT Infrastructure;\\ Transportation Networks and Logistics Management}  \\ 
    \midrule
    \makecell[l]{Social Policy\\Administration} & \makecell[l]{Education and Scientific Research; \\Government Administration and Public Sector Services;\\ Healthcare Systems and Public Health; \\Demographics and Social Development}  \\ 
    \midrule
    \makecell[l]{Consumer Lifestyle} & \makecell[l]{Retail Trade and E-commerce Platforms; \\Tourism and Hospitality Services; \\Food and Beverage Services;\\ Business Management}  \\ 
    \midrule
    Financial Economics & \makecell[l]{Economic Development and International Trade; \\Banking and Financial Services}  \\ 
    \bottomrule
    \end{tabular}}
    \captionof{table}{The 6 domains and 19 sub-domains in T2R-bench}
    \label{Tab:domains}
\end{table}

\subsection{Data Source of T2R-bench}
\label{Appd:source}

The sources of tabular data in T2R-bench are shown in Table~\ref{Tab:source}.

\begin{table}[ht]
    \centering
    \resizebox{\linewidth}{!}{
    \begin{tabular}{l l}
    \toprule
     \textbf{Sources} & \textbf{Websites} \\
     \midrule
     \multicolumn{2}{l}{\textbf{Open-source data platform}} \\
     \midrule
     Wolrd Bank Group & https://datacatalog.worldbank.org/  \\
     \midrule
     National Bureau of Statistics of China & https://www.stats.gov.cn/sj/ \\ 
     \midrule
     Kaggle & https://www.kaggle.com/datasets  \\
     \midrule
     \makecell[l]{China Association of Automobile\\ Manufactures} & http://www.caam.org.cn/  \\
     \midrule
     Beijing Public Data Open Platform & https://data.beijing.gov.cn/  \\
     \midrule
     \makecell[l]{The United States Government’s \\Open Data Site} & https://catalog.data.gov/dataset  \\
     \midrule
     \makecell[l]{China Securities Regulatory \\Commission Data Platform} & http://www.csrc.gov.cn/csrc/tjsj/index.shtml  \\
     \midrule
     Shanghai Public Data Open Platform & https://data.sh.gov.cn/view/data-resource/index.html  \\
     \midrule
     CelesTrak & https://celestrak.org/  \\
     \midrule
     \multicolumn{2}{l}{\textbf{Tabular dataset}} \\
     \midrule
     MiMoTable\cite{li2024mimotable} & https://github.com/jasonNLP/MiMoTable  \\
     \bottomrule
    \end{tabular}}
    \caption{The data sources of T2R-bench Tables}
    \label{Tab:source}
\end{table}

\section{Implementation Details for Evaluation Criteria}
\subsection{Prompts for Numerical Accuracy Criterion}
\label{Appd:NAC}

This subsection introduce the details of evaluating numerical accuracy criterion. Firstly, given the report to be evaluate, we extract clusters of sentences with numerical values through using regular expressions. Secondly, we transfer the extracted sentence clusters with numerical statements to inversely generate questions which take these sentence clusters as answers, following the prompt below:

\begin{tcolorbox}[sidebyside, sidebyside align=top seam, width=\linewidth, colback=gray!20, colframe=white, colbacktitle=white, coltitle=white, breakable, arc=0mm, left=0mm, right=0mm]
\small
As an expert in language logic analysis and data recognition, you need to transform the given sentence into question addressing the numerical parts. The questions should inquire about all numerical values appearing in the paragraph, clearly specifying the objects and criteria based on the context.\\

\#\# Example:\\

\hspace{15pt}
Input: In July 2023, the Kangming Road 4S store sold 20 Accord, 116 Odyssey, 35 Vezel, 123 CR-V, 43 Lingpai, 163 Fit, and 39 Odyssey units.\\

\hspace{15pt}
Output: How many units of Accord, Odyssey, Vezel, CR-V, Lingpai, Fit, and Odyssey were sold by the Kangming Road 4S store in July 2023?\\

Input Sentence:\\
{\red [SENTENCE]}\\

Please directly output the question, do not include any additional explanations or
comments.
\end{tcolorbox}

Thirdly, we get the answer of each question by prompting three different LLMs' coder versions (Qwen2.5-32B-Coder-Instruct, Deepseek-Coder and CodeLlama-70B-Instruct) to generate python code and extract relative data through Python programming, following the ideas of previous research proposed for Table QA task. If the code execution fails, it will not be included in the final score. The code generation prompt is shown below:

\begin{tcolorbox}[sidebyside, sidebyside align=top seam, width=\linewidth, colback=gray!20, colframe=white, colbacktitle=white, coltitle=white, breakable, arc=0mm, left=0mm, right=0mm]
\small
You are a data analysis assistant. Based on the user's provided analysis question, analytical approach, and file path, generate an efficient and robust Python code snippet to read the file from the specified path and perform data extraction.\\

\#\# Data Description (Input):\\
{\red [DATA]}\\

\#\# Specified File Path:\\
{\red [FILE PATH]}\\

\#\# User Query:\\
{\red [QUERY]}\\

\#\# Requirements:\\

File Reading: \\

\hspace{15pt}
Efficiently read data from the specified path according to the file format and size (supporting CSV, Excel, etc.) and load it into a pandas DataFrame. For larger datasets, choose appropriate reading methods to ensure performance.\\

Data Processing:\\

\hspace{15pt}1. Process data based on the user's requirements and analytical approach, including column selection, conditional filtering, group calculations, etc., ensuring the code results meet the user's query. All computed results should retain two decimal places for precise representation.

\hspace{15pt}2. Analyze the user's query and the execution process of the Python code in the Chain-of-Thought (COT) manner.

\hspace{15pt}3. Limit the number of keys in the output dictionary to within 10, avoiding tuple data types in the output.

\hspace{15pt}4. Avoid using object types and numpy operations; ensure correct computation types during calculations.

\hspace{15pt}5. Ensure all keys and values in the answer conform to dictionary format requirements, with keys being string types and values being strings or dictionaries, not lists or tuples, and convert types as necessary.

\hspace{15pt}6. The generated code should be robust, including error handling and file format compatibility. It should strictly match column names mentioned in the user's query, avoiding irrelevant or mismatched columns.

\hspace{15pt}7. Return variable format: The final result should only include the `answer` variable in dictionary format, without any other outputs.\\

\#\# Example:\\
{\red [PYTHON CODE EXAMPLES]}\\

\#\# Note:
Ensure outputs are formatted compactly and effectively, allowing successful loading by Python scripts, and avoid explanatory content.\\
\end{tcolorbox}

After obtaining the three sets of answers from Qwen2.5-32B-Coder-Instruct, Deepseek-Coder, and CodeLlama-70B-Instruct, we apply a majority-voting mechanism to aggregate these outputs into the single most reliable result, using the prompt below:
\begin{tcolorbox}[sidebyside, sidebyside align=top seam, width=\linewidth, colback=gray!20, colframe=white, colbacktitle=white, coltitle=white, breakable, arc=0mm, left=0mm, right=0mm]
\small
You are a model evaluator who rigorously applies consistency based assessment principles. You excel at analyzing, synthesizing, and summarizing multiple large language model outputs, and under a majority voting scheme to deriving the most coherent and reliable final answer.\\

\#\# Task Data Description (Input):\\
{\red [ANSWER 1], [ANSWER 2], [ANSWER 3]}\\

\#\# Requirements:\\

Answer Grouping:\\

\hspace{15pt}1. Extract the core response from each model's output, then categorize the model's answers into groups of equivalent meaning. Ensure that stylistic or formatting differences (e.g., "the RPN range is 24 to 24" versus "24.0–24.0") do not lead to separate groupings when the semantic content is identical.

\hspace{15pt}2. Accurately identify semantically consistent answers to prevent fragmentation due to superficial wording variations.

\hspace{15pt}3. You may paraphrase or consolidate responses when summarizing each group.

\hspace{15pt}4. Numeric equivalence rule: If two numeric answers round to the same value, consider them identical; adopt the value with the highest precision as the representative.\\

Final Answer Determination:\\

\hspace{15pt}1.Apply a majority voting rule: select the answer supported by the greatest number of models as the final result.

\hspace{15pt}2. If there is a tie for highest votes, output "Unable to Infer" without subjective judgment.

\hspace{15pt}3.If the selected answer is empty, "nan", "0", "0.0", an empty list or empty dictionary, or otherwise non‐informative, also output "Unable to Infer".\\

\#\# Response Format:\\

\hspace{15pt}
\begin{minipage}{\dimexpr\linewidth-15pt}
The response must strictly follow JSON format, as shown below:\\
\{

    \hspace{15pt}"Answer Groups": \{
    
        \hspace{15pt}\hspace{15pt}"Answer 1": ["LLMA", "LLMB"]
        
        \hspace{15pt}\hspace{15pt}"Answer 2": ["LLMC", "LLMD"]
        
        \hspace{15pt}\hspace{15pt}...
        
    \hspace{15pt}\},

    \hspace{15pt}"Final Answer": "The consolidated answer, or 'Unable to Infer'"\\
\}\\
\end{minipage}

\#\# Example:\\
{\red [TASK ANSWER EXAMPLES]}\\

\#\# Note:
Ensure outputs are formatted compactly and effectively, fully understand the requirements.\\
\end{tcolorbox}

Finally, by comparing the derived answers with numerical statements within each sentence cluster, we obtain the final NAC score, using the prompt below:

\begin{tcolorbox}[sidebyside, sidebyside align=top seam, width=\linewidth, colback=gray!20, colframe=white, colbacktitle=white, coltitle=white, breakable, arc=0mm, left=0mm, right=0mm]
\small
As an expert in fact verification and logical analysis, your task is to compare a given factual statement against a standard factual answer to determine if there is any contradiction in their numerical data. You should provide a score between 0 and 1, where 1 indicates complete agreement and 0 indicates complete contradiction.\\

\#\# Note: Focus solely on the numerical portions of the statements. Only output the final score without any intermediate steps or explanations.\\

\#\# Response Format:
\{

    \hspace{15pt}"score": 1.0,

    \hspace{15pt}"reason": "Reason for the assigned score"
    
\}

\#\# Factual Statement to Verify:\\
{\red [SENTENCE]}\\

\#\# Standard Factual Answer:\\
{\red [ANSWER]}\\

Please ensure that your response is strictly formatted as a valid JSON object and can be directly parsed by `json.load()`. Do not include any additional characters, comments, or text outside of the JSON structure.'''
\end{tcolorbox}

\begin{table*}[!h]
    \centering
    \resizebox{0.9\linewidth}{!}{
    \begin{tabular}{ll}
      \toprule
    Aspect & Description \\
    \midrule
    Reasoning Depth & \makecell[l]{Does the report demonstrate deep and multi-layered reasoning behind its claims?\\
    Does the analysis go beyond surface-level observations to reveal underlying mechanisms or causes?}\\
    \midrule
    Human-like Style & \makecell[l]{Does the writing style of the report resemble natural human expression rather than overly structured \\or mechanical language generated by machines?\\ Do you think it even slightly resembles machine-generated content, or human written content?}\\
    \midrule
    Practicality & \makecell[l]{Are the analyses and recommendations provided in the report practically feasible?\\ Can they offer valuable references to readers?\\ Does the report demonstrate profound industry insights?}\\
    \midrule
    Content Completeness & \makecell[l]{Does the report provide a comprehensive overview of both current status and future opportunities?\\ Are there areas where the report’s depth of coverage is insufficient?\\ Where could additional data or examples strengthen the report’s coverage?}\\
    \midrule
    Logical Coherence & \makecell[l]{Is the report structured so that each point builds logically on the previous one?\\ Are there any gaps in reasoning or sudden jumps between topics?\\ Do all conclusions follow clearly from the evidence or analysis presented?}\\
      \bottomrule
    \end{tabular}}
    \caption{Evaluation Aspects for General Evaluation Criterion.}
    \label{Tab: evaluation_aspects}
\end{table*}

\subsection{Report Evaluation Aspects for General Evaluation Criterion}

Existing evaluation criteria for reports or long texts typically encompass multiple aspects, including relevance, logical coherence, clarity, human-like style, innovation, and structural rationality, when using LLMs as a judge\cite{bai2024longwriter, zheng2023judging, li2024llmasajudge}. However, some evaluation aspects, such as linguistic standardization and logical coherence, don not show significant difference across various methods. Therefore, we concentrate on those aspects that can effectively distinguish the quality of different reports, as shown in Table~\ref{Tab: evaluation_aspects}.

\subsection{Prompt for General Evaluation Criterion}
\label{Appd:GEC_prompt}

\begin{tcolorbox}[sidebyside, sidebyside align=top seam, width=\linewidth, colback=gray!20, colframe=white, colbacktitle=white, coltitle=white, breakable, arc=0mm, left=0mm, right=0mm]
\small
You are a professional large-model evaluation expert specializing in assessing the quality of AI-generated reports. We will provide you with a user instruction and the AI-generated report. Your task is to evaluate the AI-generated report according to the following evaluation criteria and scoring rules.\\

\#\# Evaluation Aspects:\\
{\red [EVALUATION ASPECTS]}\\

\#\# Scoring Criteria:\\

\hspace{15pt}
\begin{minipage}{\dimexpr\linewidth-15pt}
The score ranges from 0 to 10. The intermediate ranges are defined as follows:\\
- 10 points: Fully meets requirements, outstanding performance, comprehensive content, no obvious defects.\\
- 8–9 points: Strong performance, meets most requirements with only minor flaws, very close to perfect overall.\\
- 6–7 points: Some shortcomings or areas that need improvement, yet still generally meets the requirements and provides valuable information or analysis.\\
- 4–5 points: Noticeable flaws or omissions; certain requirements are not adequately addressed, negatively affecting overall quality.\\
- 0–3 points: Poor quality; fails to satisfy core requirements. Contains serious errors, omissions, or logical confusion that prevent effective communication of information.\\
\end{minipage}

\#\# Response Format:\\

\hspace{15pt}
\begin{minipage}{\dimexpr\linewidth-15pt}
The response must strictly follow JSON format, as shown below:\\
\{

    \hspace{15pt}"Evaluation Aspect 1": \{
    
        \hspace{15pt}\hspace{15pt}"Reason": "Explanation for this aspect’s rating",
        
        \hspace{15pt}\hspace{15pt}"Score": <numeric score>
        
    \hspace{15pt}\},
    
    \hspace{15pt}...
    
    \hspace{15pt}"Evaluation Aspect N": \{
    
        \hspace{15pt}\hspace{15pt}"Reason": "Explanation for this aspect’s rating",
        
        \hspace{15pt}\hspace{15pt}"Score": <numeric score>
        
    \hspace{15pt}\}\\
\}\\
\end{minipage}

\#\# Evaluation Steps:\\

\hspace{15pt}
\begin{minipage}{\dimexpr\linewidth-15pt}
1.Understand the User Question: Carefully read the user’s request. Identify each requirement and how they interrelate.\\
2.Analyze the AI-Generated Report: Thoroughly review the report to ensure you understand its content and topic.\\
3.Evaluate Each Dimension: Check the report against every dimension in the list. Your evaluation should be both strict and fair, to enable comparison across different models.\\
4.Assign Scores and Provide Explanations: Give each dimension a score (0–10) and clearly state the reasons that justify your score.\\
5.Output the Final Evaluation: Present your results in JSON format, double-checking for any formatting or syntax errors.\\
\end{minipage}

\#\# Example:\\
{\red [EXAMPLES]}\\

\#\# Input User Question:\\
{\red [QUESTION]}\\

\#\# Input AI-Generated Report:\\
{\red [REPORT]}\\

Please ensure that your response is strictly formatted as a valid JSON object and can be directly parsed by `json.load()`. Do not include any additional explanations, comments, or extraneous characters outside of the JSON structure.
\end{tcolorbox}

\section{Implementation Details for Experiments}

\subsection{Prompt Template for Generating Report by LLMs}
\label{Appd:prompt_report}

\begin{tcolorbox}[sidebyside, sidebyside align=top seam, width=\linewidth, colback=gray!20, colframe=white, colbacktitle=white, coltitle=white, breakable, arc=0mm, left=0mm, right=0mm]
\small
As an expert with extensive experience in data analysis and report writing, Please generate a comprehensive report based on the provided data and analysis perspectives, and follow these guidelines:\\

\#\# Report Standards:\\

\hspace{15pt}
\begin{minipage}{\dimexpr\linewidth-15pt}
Objectivity: Ensure that the analysis is grounded in the actual data provided by the user. Avoid subjective judgments and ensure accuracy.\\
Precision: Each conclusion should be supported by data. Ensure numbers and results are accurate and based solely on the data provided.\\
Logic: The structure of the analysis should be clear and logically connected from problem definition to conclusions and recommendations.\\
Readability: Present the analysis in a simple and straightforward manner, avoiding overly complex terminology for better understanding by non-specialists.\\
Action-Oriented: Beyond just reviewing the data, provide specific suggestions or strategies to support decision-making.\\
Variety and Pacing: Use varied language to maintain reader interest and enhance the professionalism and appeal of the analysis.\\
\end{minipage}

\#\# Requirements:\\

\hspace{15pt}
\begin{minipage}{\dimexpr\linewidth-15pt}
1. The output report should be complete and well-structured, with a minimum length of 1000 words.\\
2. Ensure content is appropriately detailed, avoiding repetition and vague descriptions.\\
3. Adjust the analysis perspective if there are gaps or incomplete data, rather than mentioning "insufficient data."\\
4. Follow the analysis standards closely, avoid directly applying template references, and tailor the content according to the actual data for proper derivation and summary.\\
\end{minipage}

\#\# Input Question for generating report:\\
{\red [QUESTION]}\\

\#\# Input Table Data for generating report:\\
{\red [TABLE DATA]}\\
\\
Please directly output the generated report, do not include any additional explanations or comments.
\end{tcolorbox}

\begin{figure*}[ht!]
\centering 
\includegraphics[width=0.95\textwidth]{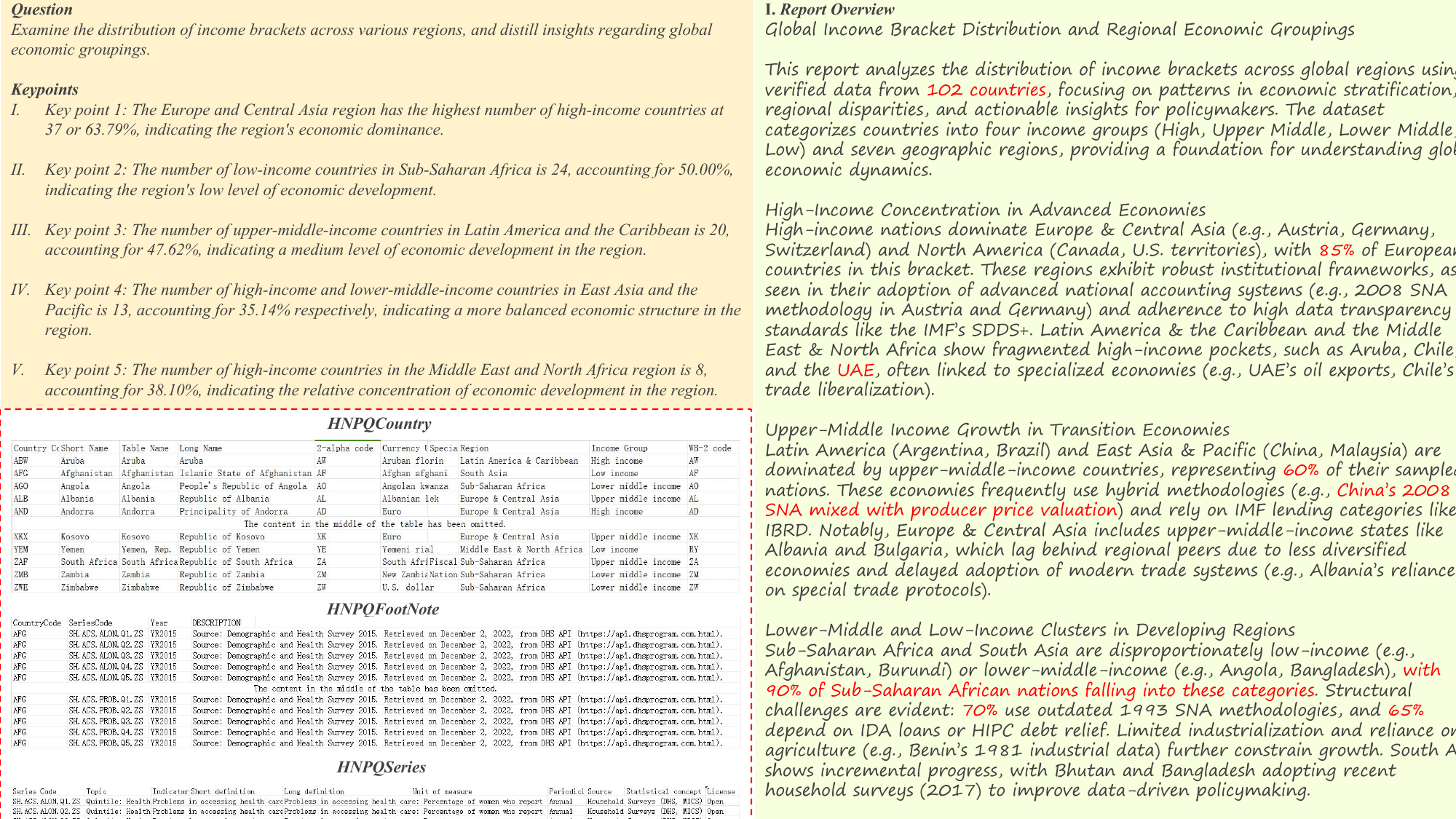} 
\caption{Case study of English extremely large-size table}
\label{Fig:case_large}
\end{figure*}

\begin{figure*}[ht!]
\centering 
\includegraphics[width=0.95\textwidth]{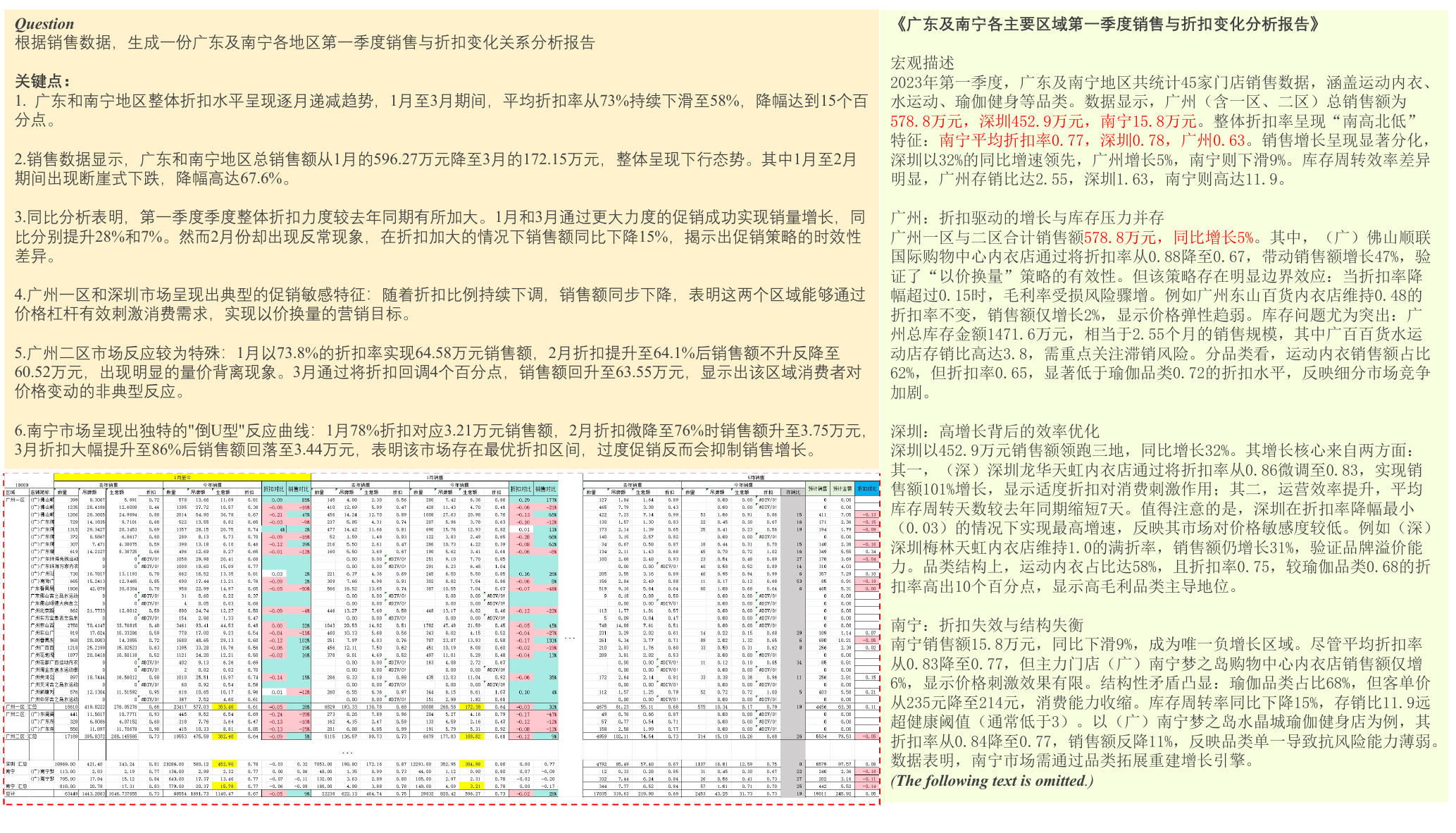} 
\caption{Case study of Chinese complex structured table}
\label{Fig:case_complex}
\end{figure*}

\subsection{Analysis of Detailed Case Study} 
\label{Appd:Case Analysis}
This subsection shows examples of a $<$question, table, report keypoints, case study$>$ combination to display detailed case study , with report generated through the single LLM of Deepseek-R1 (i.e., it is the best-performing model in our benchmark.). The incorrect parts in the report have been highlighted in red. Since the generated report is quite lengthy, the part of report has been omitted, and only the content that requires case analysis is displayed.
\\
\textbf{Case Study of English Extremely Large-size Table.}
As indicated by the highlighted text in red in Figure~\ref{Fig:case_large}, the number of countries in the sentence "this report analyzes the distribution of income brackets across global regions using verified data from 102 countries" is indeed 217. This may be due to a truncation when inputting extremely large tabular data into LLM. Moreover, the second paragraph doesn't cover Keypoint 4 when analyzing high income concentration and be lack of correct supporting data. This directly reduces the ICC evaluation metric of the report.
\\
\textbf{Case Study of Chinese Complex Structured Table.} 
For the aforementioned complex structured table shown in Figure~\ref{Fig:case_complex}, which features a complicated header and describes the comparison of sales between this year and last year from January to May, there have been numerous numerical hallucinations and incorrect conclusions. For example:
"Data shows that Guangzhou (including Zone 1 and Zone 2) had total sales of ¥5.788 million, Shenzhen ¥4.529 million, and Nanning ¥158,000. The overall discount rates exhibited a 'higher in the south, lower in the north' pattern: Nanning's average discount rate was 0.77, Shenzhen 0.78, and Guangzhou 0.63."
Here, Shenzhen’s total sales figure was taken from the "January to May cumulative data" rather than the summation of the first quarter’s, and the discount rates were also incorrect. These errors, along with challenges posed by complex table structures, descriptive hallucinations, and variable misinterpretations, reveal fundamental reasoning limitations.

\subsection{Error Analysis of Samples}
\label{Appd:Error Analysis}
As described in the case study section, we conduct an error analysis by randomly selecting 50 samples (with each set of 10 samples representing the typical characteristics of a specific table type). 

\begin{table}[!ht]
    \centering
    \resizebox{\linewidth}{!}{
    \begin{tabular}{c c c}
    \toprule 
    Error Types & Affected criteria & Ratio\\  
    \midrule
    Numerical Factual Errors & NAC & 22\% \\ 
    \midrule
     Table Structure \\
     Understanding Errors & NAC, ICC & 16\% \\ 
    \midrule
    Missing key points & ICC & 17\%  \\
    \midrule
     Generation Errors & NAC, ICC & 20\%  \\
    \midrule
     Truncation Errors & NAC, ICC & 25\%  \\
    \bottomrule
    \end{tabular}}
    \caption{Error type distribution}
    \label{table:error_analysis}
\end{table}

The primary error types identified are as follows: First, there are \textbf{hallucination errors}, which include numerical factual errors (such as incorrect numerical calculations or hallucinations of numbers from the table in the report), generation errors (such as generating content unrelated to the table, or producing incorrect or insufficiently supported conclusions or descriptions), and table structure understanding errors (e.g., column selection errors resulting from misinterpretation of table structures, such as selecting wrong column names due to incorrect recognition of complex table headers and structures; cross-table selection errors where the model retrieves data from incorrect tables). Second, there is the issue of \textbf{missing key information}, where the generated reports do not fully cover the key points, directly leading to a low ICC evaluation metric. Third, there are columns \textbf{truncation errors} when the table content exceeds the context window length (e.g., for an extremely large-size table, miscalculating a column's mean value), which directly results in a low NAC evaluation metric. The statistics of the sampling error analysis are shown in the Table~\ref{table:error_analysis}.



\subsection{URLs of Closed-source Models}

\label{Appd:websites}
\begin{table}[ht!]
   \centering
    \resizebox{0.7\linewidth}{!}{
    \begin{tabular}{ll}
    \toprule
    \textbf{Model} & \textbf{URL} \\ 
    \midrule
    \makecell[l]{Moonshot-V1-32k} & \makecell[l]{https://kimi.moonshot.cn}  \\ 
    \midrule
    \makecell[l]{Claude-3.5-Sonnet} & \makecell[l]{https://www.anthropic.comt}  \\ 
    \midrule
    \makecell[l]{Doubao-Pro-128k} & \makecell[l]{https://www.volcengine.com} \\ 
    \midrule
    \makecell[l]{Doubao-Pro-32k} & \makecell[l]{https://www.volcengine.com}  \\ 
    \midrule
    \makecell[l]{GPT-4o} & \makecell[l]{https://openai.com}  \\ 
    \midrule
    \makecell[l]{OepnAI o1-mini} & \makecell[l]{https://openai.com}  \\ 
    \bottomrule
    \end{tabular}}
    \captionof{table}{The URLs of closed-source models we used}
    \label{Tab:api}
\end{table}

\subsection{Analysis of Input Formatting}
\label{Appd:Input_Formatting}

Table~\ref{Tab:input_format} demonstrates that among the three most representative table input formats (Markdown, HTML, and JSON), the Markdown format achieves the highest average performance, followed by HTML, while JSON exhibits the lowest performance. 
\begin{table}[!ht]
    \centering
    \resizebox{0.9\linewidth}{!}{
    \begin{tabular}{cccc}
    \toprule
    ~ & Markdown & JSON & HTML \\
    \midrule
    Qwen2.5-72B-Instruct & 59.59 & 55.82 & 54.91 \\
    Deepseek-R1 & 62.71 & 58.12 & 60.02 \\ 
    OpenAI-o1-1217 & 62.76 & 59.43 & 59.67 \\ 
    \bottomrule
    \end{tabular}}
    \caption{Average performance of NAC, ICC and GEC of three different models across different input formats.}
    \label{Tab:input_format}
\end{table}

\section{Details for payment and GPU hours}
We pay each annotator a daily remuneration of \$40. We paid a total of \$2500 for calling various LLMs API interfaces. We use 16 A100 40G GPUs for inference, which took a total of 25 hours.

\end{document}